\documentclass[10pt,conference]{IEEEtran}
\usepackage{amsmath,amsfonts}
\usepackage{array}
\usepackage[caption=false,font=normalsize,labelfont=sf,textfont=sf]{subfig}
\usepackage{textcomp}
\usepackage{stfloats}
\usepackage{url}
\usepackage{verbatim}
\usepackage{graphicx}
\usepackage{cite}
\usepackage[ruled,vlined,linesnumbered]{algorithm2e}
\usepackage{algpseudocode}
\usepackage{xcolor}
\usepackage{multirow}
\usepackage{amsfonts}
\usepackage{xspace}
\usepackage{colortbl}
\usepackage{makecell}
\usepackage{microtype}
\usepackage{enumitem}
\usepackage{pifont}
\usepackage{booktabs}
\usepackage{amsthm, amssymb}
\newtheorem{definition}{Definition}
\newcommand{\tool}{\textit{Decictor}\xspace}

\newcommand{\cmf}[1]{{\textcolor{black}{#1}}} %\textcolor{black}
\newcommand{\todo}[1]{{\textcolor{orange}{todo:#1}}}

\newcommand{\ods}{ODS}
\newcommand{\nods}{NoDS}
\newcommand{\randombasline}{{Random}\xspace}
\newcommand{\randomM}{{Random-$\delta$}\xspace}

\newcommand{\route}{path\xspace}

\DeclareMathOperator*{\argmax}{arg\,max}

\usepackage{xpatch}
\makeatletter
\patchcmd\@makecaption{\\}{.~}{}{\fail}
\begin{document}
% \linenumbers 
% \title{\tool: Detecting Non-Optimal Decisions in Autonomous Driving Systems via Metamorphic Testing}
% \title{\tool: Detecting Non-Optimal Decisions in Autonomous Driving Systems}
% \title{\tool: Towards Evaluating the Robustness of \cmf{Path-Planning Decision} in Autonomous Driving Systems}
\title{\tool: Towards Evaluating the Robustness of Decision-Making in Autonomous Driving Systems}
% \author{Anonymous Author(s)*}
% \author{Mingfei Cheng, Yuan Zhou, Xiaofei Xie, Junjie Wang, Guozhu Meng, and Kairui Yang
% %
% \thanks{Mingfei Cheng and Xiaofei Xie are with the School of Computing and Information Systems (SCIS), Singapore Management University, Singapore.\\
% E-mail: mfcheng.2022@phdcs.smu.edu.sg, xfxie@smu.edu.sg
% }% 
% \thanks{Yuan Zhou is with the School of Computer Science and Engineering, Nanyang Technology University, Singapore.\\
% E-mail: y.zhou@ntu.edu.sg}
% \thanks{Junjie Wang is with the College of Intelligence and Computing, Tianjin University, China.\\
% E-mail: junjie.wang@tju.edu.cn}
% \thanks{Guozhu Meng is with the Institute of Information Engineering, Chinese Academy of Sciences, China.\\
% E-mail: mengguozhu@iie.ac.cn}
% \thanks{Kairui Yang is with the DAMO Academy, Alibaba Group, China.\\
% E-mail: kairui.ykr@alibaba-inc.com}
% }

\author{
    \IEEEauthorblockN{Mingfei Cheng$^{1}$, Xiaofei Xie$^{1}$, Yuan Zhou$^{2}\IEEEauthorrefmark{1}$, Junjie Wang$^{3}$, Guozhu Meng$^{4}$ and Kairui Yang$^{5}$}
    \IEEEauthorblockA{$^1$Singapore Management University, Singapore $^2$Zhejiang Sci-Tech University, China}
    \IEEEauthorblockA{$^3$Tianjin University, China $^4$Chinese Academy of Sciences, China $^5$Alibaba Group, China}
    % DAMO Academy, 
    % \IEEEauthorblockA{}
    % \IEEEauthorblockA{}
    % \IEEEauthorblockA{$^5$ DAMO Academy, Alibaba Group, China}
    % \IEEEauthorblockA{\textit{jcyu.2022@phdcs.smu.edu.sg, \{wuyuechen, chenyingfeng1, huyujing\}@corp.netease.com}} 
    % \IEEEauthorblockA{\textit{xfxie@smu.edu.sg, weile@iastate.edu, ma.lei@acm.org, fanzhang@zju.edu.cn}}
    % \thanks{\IEEEauthorrefmark{1} Corresponding author}
    % \thanks{\IEEEauthorrefmark{2} Contributed equally}
    % For author and content space
    % \IEEEauthorblockA{Email: \{mfcheng.2022, xfxie\}@smu.edu.sg, yuanzhou@zstu.edu.cn}
    % \IEEEauthorblockA{junjie.wang@tju.edu.cn, mengguozhu@iie.ac.cn, kairui.ykr@alibaba-inc.com}
    \vspace{-2.1em} 
}

% \markboth{Journal of \LaTeX\ Class Files,~Vol.~18, No.~9, September~2020}{}

% {How to Use the IEEEtran \LaTeX \ Templates}

\maketitle

\begingroup\renewcommand\thefootnote{}\footnotetext{\IEEEauthorrefmark{1} Corresponding author}\endgroup

\thispagestyle{plain}
\pagestyle{plain}

\begin{abstract}
Autonomous Driving System (ADS) testing is crucial in ADS development, with the current primary focus being on safety. However, the evaluation of non-safety-critical performance, particularly the ADS's ability to make optimal decisions and produce optimal paths for autonomous vehicles (AVs), is also vital to ensure the intelligence and reduce risks of AVs. 
Currently, there is little work dedicated to assessing \cmf{the robustness of ADSs' path-planning decisions (PPDs), i.e., whether an ADS can maintain the optimal PPD after an insignificant change in the environment.} The key challenges include the lack of clear oracles for assessing \cmf{PPD} optimality and the difficulty in searching for scenarios that lead to non-optimal \cmf{PPDs}. To fill this gap, in this paper, we focus on evaluating \cmf{the robustness of ADSs' PPDs} and propose the first method, \tool, for generating non-optimal decision scenarios ({\nods}s), where the ADS does not plan optimal paths for AVs. 
\tool comprises three main components: Non-invasive Mutation, Consistency Check, and Feedback. 
To overcome the oracle challenge, Non-invasive Mutation is devised to implement conservative modifications, ensuring the preservation of the original optimal path in the mutated scenarios. 
Subsequently, the Consistency Check is applied to determine the presence of non-optimal \cmf{PPDs} by comparing the driving paths in the original and mutated scenarios. 
To deal with the challenge of large environment space, we design Feedback metrics that integrate spatial and temporal dimensions of the AV's movement. These metrics are crucial for effectively steering the generation of {\nods}s. 
\cmf{Therefore, \tool can generate {\nods}s by generating new scenarios and then identifying {\nods}s in the new scenarios.}
We evaluate \tool on Baidu Apollo, an open-source and production-grade ADS. The experimental results validate the effectiveness of \tool in detecting non-optimal \cmf{PPDs} of ADSs. It generates \cmf{63.9} NoDSs in total, while the best-performing baseline only detects \cmf{35.4} NoDSs. 
% Our work provides valuable and original insights into evaluating the non-safety-critical performance (e.g., decision robustness) of ADSs. 
\end{abstract}

% \begin{IEEEkeywords}
% Non-optimal Decision, Autonomous Driving System, Decision Robustness Testing
% \end{IEEEkeywords}

\section{Introduction}
% Autonomous Vehicles (AVs) are highly successful in building smart cities with the advancement of autonomous driving systems (ADSs).

Autonomous Driving Systems (ADSs) have been a revolutionary technology with the potential to transform our transportation system into an intelligent one.
ADSs aim to enable vehicles to operate without human intervention, relying on a combination of different sensors (e.g., camera, radar, lidar, and GPS) and artificial intelligence algorithms to perceive the environment, make decisions, and navigate safely.
Even though the development of ADSs has seen significant progress over the past few decades, it is still a great challenge to guarantee that the ADSs can satisfy all performance requirements under different situations due to the existing vulnerabilities in ADSs~\cite{garcia2020comprehensive}. 
Therefore, before their real-world deployment, ADSs should be sufficiently tested  in all aspects~\cite{lou2022testing}. 

\begin{figure}[!t]
    \centering
    \includegraphics[width=\linewidth]{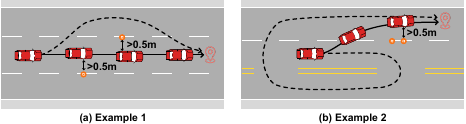}
    % \vspace{-3mm}
    \vspace{-20pt}
    \caption{\cmf{Illustration of non-optimal PPDs. \textbf{Solid lines} are expected optimal paths. \textbf{Dotted lines} are non-optimal paths planned by the ADS.}}
    \label{fig:intro}
    % \vspace{-5mm}
    \vspace{-15pt}
\end{figure}

Usually, the requirements of ADSs can be classified as safety-critical and non-safety-critical. Safety-critical requirements are those essential for ensuring safe operations and performance of the autonomous vehicle.
% directly related to vehicle and environmental safety.
For example, an ADS should guarantee that the ego vehicle (i.e., the autonomous vehicle controlled by the ADS) should arrive at its destination without causing collisions or violating traffic rules. 
Non-safety-critical requirements refer to aspects that do not directly impact safety but are important for user experiences and optimal vehicle performance, such as motion efficiency, passenger comfort, and energy consumption.

Currently, various testing technologies have been proposed to evaluate the safety-critical requirements of ADSs (referred to as ``safety testing''), such as data-driven methods~\cite{zhang2023building,deng2022scenario,Bashetty20DeepCrashTest,paardekooper2019automatic} and \cmf{search-based methods~\cite{gambi2019automatically, av_fuzzer,cheng2023behavexplor,icse_samota,tse_adfuzz,aleti2022identifying,css_drivefuzzer,huai2023sceno,huai2023doppelganger,tian2022mosat,lu2022learning}}.
They aim to generate safety-critical scenarios, under which the ego vehicle will cause safety issues, such as collisions and traffic rule violations. \textit{To the best of our knowledge, there is limited research on evaluating the robustness of the \cmf{path-planning decisions (PPDs)} in ADSs, i.e., their ability to plan optimal paths in dynamic environments}.
\cmf{PPD optimality} is an important non-safety-critical property for guaranteeing the motion efficiency of the ego vehicle in complex traffic scenarios, but it may not be achieved by ADSs. 
\cmf{Fig.~\ref{fig:intro} presents two illustrative examples that highlight PPD robustness issues, where the ADS fails to determine the optimal paths 
% due to inaccurate reasoning 
after minor environmental changes, 
% such as the addition of two obstacles 
such as placing two small obstacles on the lane boundaries 
that would not affect the optimal path (i.e., solid lines). Note that, in these changes, the ego vehicle still maintains a safe distance~\cite{FINAL} (i.e., more than 0.5 meters from the obstacle) to pass the scenarios if it follows the optimal paths.} The video \cite{teslaexample} shows an example that Tesla FSD cannot make robust \cmf{PPDs} at a fork in a real-world road.

An ADS with limited intelligence might inaccurately assess traffic scenarios, resulting in suboptimal choices. Such decisions could cause prolonged delays or select inefficient routes, leading to unnecessary hold-ups or even elevating safety risks. Therefore, evaluating the \cmf{robustness of PPD} of ADSs is critically vital, as it affects the comfort and efficiency of autonomous vehicles and could pose additional safety concerns.
{Note that, in this paper, \cmf{PPD} robustness evaluation aims to explore scenarios where safety issues are not present, instead focusing on assessing the optimal \cmf{PPD} capabilities of ADS within dynamic driving environments.}

Technically, drawing on the foundational principles established by existing research in robustness evaluation research~\cite{Goodfellow2015,cw2017}, the core objective of robustness evaluation is to assess a model's capacity to maintain correct prediction following some perturbations. For instance, in image classification, an image is slightly modified (e.g., through the addition of noise) to test whether the model can still accurately classify the perturbed image. Gradient-based methods~\cite{Goodfellow2015,cw2017} are commonly employed to efficiently calculate these perturbations. However, evaluating the \cmf{PPD} robustness of ADS presents unique challenges: \ding{182} Contrary to safety testing, which relies on explicit, measurable criteria (like collision detection or timeout occurrences) to identify safety-critical violations, the assessment of \cmf{PPD} optimality in ADS under new scenarios lacks clear oracles. While traditional robustness evaluations might control the extent of perturbations (e.g., the level of noise) to ensure the perturbed inputs match the truth labels of original inputs, the complexity inherent in AV scenarios complicates the control of scenario perturbations. \ding{183} The unpredictable and infinitely variable nature of driving conditions, influenced by diverse road users and environmental factors, complicates the search of ``\textbf{\underline{N}}on-\textbf{\underline{o}}ptimal \textbf{\underline{D}}ecision \textbf{\underline{S}}cenarios'' ({\nods}s). The conventional gradient-based method is difficult to be applied in these scenarios due to discontinuities in the input space and the logical constraints that perturbations must obey, such as adherence to physical laws and ensuring that obstacles remain within realistic bounds. \ding{184} The evaluation of robustness typically requires a method to check prediction consistency between the original input and its perturbed counterpart. In classification tasks, the consistency check is straightforwardly verified by comparing their predicted labels. However, in ADS, decisions are represented by planned paths, which introduces complexity into the process of determining their consistency.

To address these challenges, we propose a novel testing method, called \tool, comprising three main components \textit{Non-invasive Mutation}, \textit{Feedback} and \textit{Consistency Check}. The innovative approach enables the effective and efficient identification of non-optimal \cmf{PPDs} and their associated NoDSs. Specifically, to address the first challenge \ding{182}, we design a \textit{non-invasive mutation} that conservatively modifies the behaviors of other participants within the driving environment in a way that the modifications would not affect the originally optimal path in the given ``\textbf{\underline{O}}ptimal \textbf{\underline{D}}ecision \textbf{\underline{S}}cenario'' (ODS).  For the second challenge \ding{183}, we propose a fitness function designed to guide the generation of NoDSs. The function scores are based on two key metrics: firstly, it seeks to maximize the discrepancy in the driving paths between the ego vehicle in both the mutated and original scenarios, directly accentuating the non-optimal behaviors. Secondly, it assesses the variances between the scenarios from a broader perspective of ego behavior, measured by a vector encompassing velocity, acceleration, and orientation. The second metric is crucial as the behavior of the ego vehicle may directly affect its driving path. For the third challenge \ding{184}, given the possible slight variations between the new driving path (in the mutated scenario) and the initially optimal path (in ODS), the direct path-level comparison is not reliable. To resolve this, we introduce an abstraction-based behavior comparison method. This technique evaluates the driving patterns in both scenarios through abstraction, where significant differences in routes indicate deviations from optimal \cmf{PPDs} by the ADS.

We have evaluated \tool on the Baidu Apollo with its built-in SimControl~\cite{apollo}. We compared \tool with \cmf{nine} baseline methods: two random methods, i.e., \randombasline, which generates scenarios randomly, and \randomM, which applies our non-invasive mutation and random selection to generate scenarios, \cmf{along with seven state-of-the-art ADS testing methods, i.e., AVFuzzer~\cite{av_fuzzer}, SAMOTA~\cite{icse_samota}, BehAVExplor~\cite{cheng2023behavexplor}, DriveFuzz~\cite{css_drivefuzzer}, scenoRITA~\cite{huai2023sceno}, DoppelTest~\cite{huai2023doppelganger} and DeepCollision~\cite{lu2022learning}.} The evaluation results reveal the effectiveness and efficiency of \tool in detecting NoDSs, demonstrating the \cmf{path-planning} robustness issues within the ADS.
For example, \tool generates a total of \cmf{63.9} NoDSs
while the best-performing baseline only detects \cmf{35.4} NoDSs. 
Further experimental results demonstrate the usefulness of our mutation operation and feedback mechanisms meticulously designed within \tool. 
% Moreover, the comparison results emphasize the necessity of \tool for optimal decision testing.
% is significantly more effective and efficient at detecting NoDSs than sate-of-the-art ADS testing approaches (\todo{[]] vs. []}), i.e., []. 
% We also observed that \tool is capable of identifying safety-critical violations ([]), such as task failures, that can be prevented by adhering to the original optimal decision. 
% We also observed that state-of-the-art methods detected much less NoDSs, with 1.4, 1.6, and 5.4 for AVFuzzer, SAMOTA, and BehAVExplore respectively. The primary reason is that their optimization focus on safety violations rather than non-optimal decisions. }

% \note{Reviews: unfair comparison with these methods. (delete the description of baselines)}

In summary, this paper makes the following contributions:
\begin{enumerate}[leftmargin=*]
\item We are the first to investigate the problem of \cmf{non-optimal PPDs} made by ADSs. Our work provides valuable and original insights into evaluating the non-safety-critical performance of ADSs. 

\item We develop a search-based method that efficiently generates NoDSs and evaluates the \cmf{PPD} robustness of ADSs. 

\item We conduct extensive experiments to evaluate the effectiveness and usefulness of \tool on Baidu Apollo. As a result, a total of \cmf{63.9} NoDSs is discovered on \cmf{six initial ODSs} on average, demonstrating the potential and value of \tool in detecting \cmf{non-optimal PPDs} made by ADSs.
\end{enumerate}

\section{Background and Notation}
\subsection{Autonomous Driving Systems} \label{sec-ads-spec}
% \noindent \textbf{Autonomous Driving Systems (ADSs).} , i.e., the ego vehicles
ADSs control the motion of autonomous vehicles. Existing ADSs mainly contain two categories: End-to-End (E2E) systems \cite{roach_iccv, hu2023_uniad, openpilot}, and module-based ADSs \cite{apollo, autoware, gog2021pylot}. E2E systems use united deep learning models to generate control decisions from sensor data directly. Recently, the rapid development of Deep Learning have led to high-performance E2E systems in close-loop datasets, such as OpenPilot \cite{openpilot}. 
However, these E2E systems still perform poorly on unseen data. In contrast, module-based ADSs have better performance in various scenarios. 
% However, these E2E systems still perform poorly on unseen testing data, such as easily colliding with obstacles. In contrast, module-based ADSs have better performance in various scenarios.  and Autoware \cite{autoware}
A typical module-based ADS, such as Baidu Apollo \cite{apollo}, usually consists of localization, perception, prediction, planning, and control modules to generate decisions from rich sensor data. The localization module provides the location of the ego vehicle by fusing multiple input data from GPS, IMU, and LiDAR sensors. The perception module takes camera images, LiDAR point clouds, and Radar signals as inputs to detect the surrounding environment (e.g., traffic lights) and objects (e.g. other vehicles) by mainly using deep neural networks. The prediction module is responsible for tracking and predicting the trajectories of all surrounding objects detected by the perception module. Given the results of perception and prediction modules, the planning module then generates a local collision-free trajectory for the ego vehicle. Finally, the control module converts the planned trajectory to vehicle control commands (e.g., steering, throttle, and braking) and sends them to the chassis of the vehicle. In this paper, we choose to test module-based ADSs on a simulation platform, where the ADS connects to a simulator via a communication bridge. Same to previous works \cite{cheng2023behavexplor,huai2023doppelganger,huai2023sceno}, the ADS receives perfect perception data from the simulator, and sends the control commands to the ego vehicle in the simulator.

\subsection{Scenario} \label{def-scenario}
ADS testing necessitates a collection of scenarios as inputs. 
Each scenario is characterized by a specific environment (e.g., road), the scenery and objects (e.g., static obstacles and Non-Player Character (NPC) vehicles).
Complex scenarios are generated by combining relevant attributes from the Operational Design Domains (ODDs)~\cite{thorn2018framework}. Due to the vast attribute space, covering all attributes with all possible values is impractical. \cmf{Existing studies~\cite{av_fuzzer, cheng2023behavexplor, icse_samota, huai2023sceno, huai2023doppelganger, lu2022learning, tian2022mosat, css_drivefuzzer}} address this by selecting specific subsets of attributes. In this paper, we evaluate the non-safety-critical performance of ADS motion using scenarios formulated with traffic cones as static obstacles and NPC vehicles as dynamic objects.

Therefore, a \emph{scenario} can be described as a tuple $s = \{\mathcal{A}, \mathcal{P}\}$, where $\mathcal{A}$ is the motion task of the ADS under test, including the start position and the destination, $\mathcal{P}$ is a finite set of participants, including the set of static obstacles and dynamic NPC vehicles. \cmf{Note that in our paper, participants in the scenario do not include the ego vehicle controlled by the ADS.} Scenario observation is a sequence of scenes within the execution of the scenario, and each scene represents the states of the ego vehicle and other participants at a timestamp. Formally, given a scenario $s = \{\mathcal{A}, \mathcal{P}\}$, its observation is denoted as $O(s)=\{o_0, o_1, \ldots, o_{k}\}$, where $k$ is the length of the observation and $o_i$ is a scene at timestamp $i$. In detail, $o_i=\{y_i^0, y_i^1, \ldots, y_i^{|\mathcal{P}|}\}$ where $y_i^j=\{p_{i}^{j}, \theta_{i}^{j},v_{i}^{j},a_{i}^{j}\}$ denotes the waypoint of a participant $j\in\mathcal{P}$ at timestamp $i$, including the center position $p_{i}^{j}$, the heading $\theta_{i}^{j}$, the velocity $v_{i}^{j}$ and the acceleration $a_{i}^{j}$.
A driving \route of the participant $j$ can be defined as $\tau^j(s)=\{p_0^j, \ldots, p_k^j\}$.
% $\tau^j(s)=\{y_0^j, \ldots, y_k^j\}$. 
By default, we use $\tau(s)$ to represent the driving \route of the ego vehicle in the scenario $s$.
Unless otherwise specified, the driving path in the following context refers to the path of the ego vehicle in $s$, i.e., $\tau(s)$.

\section{Overview}
% \cmf{In this section, we begin by defining the problem of detecting non-optimal decision scenarios. We then provide a high-level overview of our proposed method.}

% In this section, we begin by defining the problem of detecting the issues that the ADS selects non-optimal decisions and causes non-safety critical violations (i.e., a less efficient driving path) or in scenarios. We then provide a high-level overview of our proposed method.

\begin{figure*}[!t]
    \centering
    \includegraphics[width=0.85\textwidth]{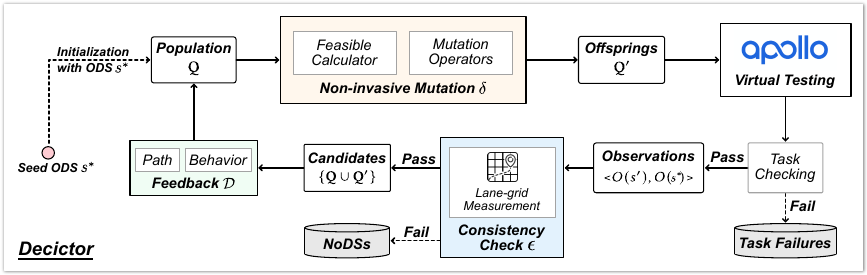}
    \vspace{-5pt}
    \caption{\cmf{Overview of \tool.}}
    \label{fig: method_overview}
    % \todo{Update s}
    \vspace{-10pt}
\end{figure*}

\subsection{Problem Definition}
This paper aims to assess the optimization capability of the \cmf{path-planning} process in ADSs. Given an ODS $s$ and its associated optimal path for the ego vehicle $\tau(s)$, \cmf{PPD} optimality testing aims to assess \cmf{the robustness of ADS's PPD}, i.e., its ability to consistently navigate along $\tau(s)$. \cmf{Following the robustness definitions of neural networks~\cite{leino2021globally, ruan2019global, papernot2016distillation, carlini2017towards}, we define the PPD robustness as follows:} 

\begin{definition} [\cmf{Path-planning Decision Robustness}] \label{def-mr}
Given an \ods $\ s=\{\mathcal{A}, \mathcal{P}\}$, we formalize the \cmf{path-planning decision (PPD)} robustness as:
\cmf{\begin{equation}\label{eq:MR} 
\forall s'.||s-s'||_\delta = True \Longrightarrow \epsilon (\tau(s), \tau(s')) = True
\end{equation}}
% where  $\delta$ represents a mutation function carefully crafted to ensure that the ego vehicle in the modified scenario $(s')$ can still navigate the \cmf{optimal driving path} from the original scenario~$(s)$. Specifically, 
where $||\cdot||_{\delta}$ is used to constrain scenario mutation, ensuring that the mutation $\delta$ modifies a subset of participants $P \subseteq \mathcal{P}$ in $s$ to produce $s'$, without compromising the optimality of the original driving path, i.e., $||s-s'||_\delta =True$. The function $\epsilon$ serves as a consistency checking mechanism, evaluating \cmf{whether the two driving paths executed by the ADS in $s$ and $s'$ are equivalent.}
\end{definition}

The definition of \cmf{PPD} robustness stipulates that, following minor perturbations (i.e., $s'=\delta(s,P)$), the ADS should maintain the capacity to select the original optimal driving path (i.e., $\pi(s)$). \textit{The main goal of the \cmf{PPD robustness} testing is to identify scenarios where the ADS can accomplish its task without encountering safety issues, yet does not manage to plan and execute the optimal driving path.} Note that the \cmf{PPD} optimality of the seed {\ods} $s$ can be manually confirmed while the \cmf{PPD} optimality of mutated scenarios \cmf{$s'$} can be automatically checked by evaluating their consistency.

\subsection{Approach Overview}
% \fei{Add overview description here}
% \todo{Framework Name.}

Fig.~\ref{fig: method_overview} provides a high-level depiction of our testing framework \tool, designed to detect non-optimal \cmf{PPDs}. The basic idea underlying our method involves generating a scenario $s'$, where the ego vehicle opts for a \route far from the optimal one in the {\ods} $s$ under the non-invasive mutation $\delta$. It can be formulated as an optimization problem aiming to maximize the difference between the driving {\route}s of $s$ and $s'$:
\begin{equation}\label{eq:obejctive}
\Delta_{s} = \argmax_\Delta \mathcal{D}(\tau(s), \tau(s')),  \text{ where }s'=\delta(s, \Delta)
\end{equation}
where $\Delta$ symbolizes the non-invasive perturbations on the scenario $s$, and $\mathcal{D}$ is a function to measure the distance or difference between the driving {\route}s of the two scenarios.
% Therefore, we n generate a NoDS from $s$: $s'=\delta(s, \Delta)$.

\tool adopts a search-based method to solve the optimization problem. The process initiates with a seed {\ods} $s^*$ and aims to output a set of violation tests, i.e., {\nods}s, considering the optimal \route in $s^*$. We first initialize a population $\mathbf{Q}$ based on ${s^*}$. \tool then optimizes this population iteratively until the testing budget $B$ exceeds.
% which comprises a certain number of new scenarios. 
% \tool then optimizes this population iteratively until it identifies at least one NoDS or the budget exceeds. 
In each iteration, \tool performs \textit{Non-invasive Mutation} ($\delta$) to generate the offspring $\mathbf{Q'}$ that has the same size as $\mathbf{Q}$. The mutation strategy ensures that the original optimal \route in $s^*$ remains optimal in the new scenarios. 
To determine if the ADS makes the optimal \cmf{PPDs} in these new scenarios, we introduce \textit{Consistency Check} ($\epsilon$), a hierarchical approach for measuring the equivalence with regard to \cmf{PPD} optimality between the seed {\ods} and these new scenarios. \textit{Consistency Check} initially monitors whether a new scenario completes the task during simulation execution. 
If the modified scenario results in a task failure, such as collisions, 
% it is categorized as a task failure and 
it is not considered as a {\nods} of interest. This is because such failures can already be detected by existing AV testing tools that utilize timeout or collision detection oracles. Conversely, if the new scenario passes the task checking, we collect the observation from the simulator. The \textit{Consistency Check} then verifies the equivalence of the driving {\route}s between the seed {\ods} and the new scenario using a lane-grid measurement. Detection of the consistency violation in this step identifies a {\nods}. If no violation occurs, \tool proceeds to select the superior population from the candidates of the old and new populations for the next iteration. This selection process is guided by the Feedback ($\mathcal{D}$) that takes into account the spatial and temporal characteristics, i.e., the driving path and the behaviors, of the ego vehicle's motion.

% \begin{figure*}[!t]
%     \centering
% 	\includegraphics[width=0.9\textwidth]{figs/decictor_mutation.pdf}
% 	\caption{Illustrative example of feasible region calculation for mutation.\todo{Add i-1}}
%  	\label{fig:mutation}
% \end{figure*}

\section{Approach}

\begin{algorithm}[!t]
\small
\SetKwInOut{Input}{Input}
\SetKwInOut{Output}{Output}
\SetKwInOut{Para}{Parameters}
\SetKwComment{Comment}{\color{blue}// }{}
\SetKwFunction{mutation}{{\textbf{NonInvasiveMutation}}}
\SetKwFunction{fitness}{{\textbf{FeedbackCalculator}}}
\SetKwFunction{test}{{\textbf{SimExecution}}}
\SetKwFunction{oracle}{{\textbf{MROracleCheck}}}
\SetKwFunction{selection}{\textbf{Selection}}

\Input{
The seed ODS: $s^*$    % Target ADS $\mathcal{A}$
}
\Output{
    The set of NoDSs: $\mathbf{F}_{n}$ 
    % Task failures $\mathbf{F}_{s}$
}
\Para{
    Population size $N$, Testing budget $B$
}

% \Fn{\MetaFuzz{$\mathbf{S}$}}{
   % $\mathbf{F}_{n} \gets \{\}$, $\mathbf{F}_{s} \gets \{\}$, $\mathbf{Q} \gets \{\}$ \\
   $\mathbf{F}_{n} \gets \{\}$, $\mathbf{Q} \gets \{\}$ \\
% \Comment*[r]{Initial failing test set}

    \For{$i \in \{1, \ldots, N\}$}{
       % $s' \gets \mutation(s)$ \\
       $\mathbf{Q} \gets \mathbf{Q} \cup \{{s}^*\}$ \\
    }

    \Repeat{Testing budget $B$ exhausted}{
        $\mathbf{Q}' \gets \{\}$\\
        \For{$\hat{s}\in \mathbf{Q}$}{
              $s' \gets \delta(\hat{s}, \Delta)$ \Comment{\color{blue}{Non-invasive Mutation}}
              $r_{tc}, O(s') \gets \test{$s'$}$ \\
              
              \If{$r_{tc}$ is passed}{
                % $\mathbf{F}_{s} \gets \mathbf{F}_{s} \cup \{s'\}$ \\
                \Comment{\color{blue}{Consistency Check}}
                % \Comment{\color{blue}{Consistency Check - Second Stage}}
                $r_{cc} \gets \epsilon(\tau(s^*), \tau(s'))$\\
                \eIf{$r_{cc}$ is passed}{
                    $\mathbf{Q'} \gets \mathbf{Q'} \cup \{s'\}$
                }{
                    $\mathbf{F}_{n} \gets \mathbf{F}_{n} \cup \{s'\}$
                }
            }
        }
        $\mathbf{Q} \gets \selection(\{\mathbf{Q} \cup \mathbf{Q'}\}, N)$  \Comment{\color{blue}{Feedback}} \label{algo0:select}
    }
    \Return $\mathbf{F}_{n}$
    % \Return \{$\mathbf{F}_{n} \cup \mathbf{F}_{s}\}$
    
\caption{\tool Algorithm}
\label{algo-metafuzz}
\end{algorithm}

Algorithm \ref{algo-metafuzz} presents the main algorithmic procedure of \tool. \tool receives a seed {\ods} $s^*$ as the input and outputs a set of {\nods}s. Parameters $N$ and $B$ can be adjusted to configure the population size and testing budget, respectively. The algorithm begins by creating an initial population with the given seed {\ods} $s^*$ (Lines 2-3). In each iteration, the algorithm first generates the offspring by mutating each scenario in the population $\mathbf{Q}$ (Lines 6-7). 
Each new scenario $s'$ is first executed 
% by the virtual testing 
in the simulation platform consisting of a simulator and the ADS under test (Line 8) and verified by the task checking $r_{tc}$ (Line 9). The new scenario passed the task checking is then evaluated for \cmf{PPD} consistency (Lines 10-14). If violated, a {\nods} is identified and added to the set $\mathbf{F}_{n}$ (Line~14). If no violation is detected, the new scenario is added to the offspring $\mathbf{Q'}$. The scenario selection picks the top $N$ scenarios from the union of $\mathbf{Q}$ and $\mathbf{Q'}$ based on their fitness scores (Line 15). The algorithm ends by returning detected {\nods}s $\mathbf{F}_{n}$ (Line 17).

In the following sections, we introduce the key components of \tool: the \textit{Non-invasive Mutation} ($\delta$), the \textit{Consistency Check} ($\epsilon$), and the \textit{Feedback} ($\mathcal{D}$).

\subsection{Non-invasive Mutation $\delta$} \label{sec:nimutation}
The main challenge of the mutation is to ensure that in the mutated scenario, the original optimal path of the ego vehicle in the seed {\ods} is not affected. 
\cmf{Following existing mutation techniques for ADS safety testing~\cite{av_fuzzer, cheng2023behavexplor, icse_samota, tang2021systematic}, we only alter the waypoints of NPC vehicles or other participants, leaving the ego vehicle unchanged.} However, the primary challenge is that these mutations could significantly impact the ego vehicle's original optimal \route.
% Besides, executing a scenario in the virtual testing is always time consuming~\cite{icse_samota}, which will reduce the efficient of detecting NoDSs. 
To mitigate this challenge, we introduce a conservative mutation approach, namely the non-invasive mutation strategy, designed to create new scenarios with less impact on the original optimal \route. While it may be theoretically difficult, or even impossible, to ensure that changes do not affect the ego vehicle's original optimal path, our method adopts a conservative approach. This involves calculating ``safe zones'' for mutation, where adjustments to participants' behaviors are unlikely to influence the original optimal path. This concept is akin to existing adversarial attack methodologies, where the difference between the new and original inputs is limited within an $L_P$ norm boundary~\cite{carlini2017towards}. Such constraints are intended to preserve the semantic integrity of the input, but a theoretical guarantee of no impact is challenging to establish~\cite{cw2017,wang2023distxplore}.
% Our experiments have also demonstrated the effectiveness of the non-invasive mutation method.

The non-invasive mutation mainly includes \textit{adding} new participants within the non-invasive feasible areas, \textit{removing} existing added participants \cmf{and \textit{changing} existing participants.}

\textbf{Non-invasive Feasible Area}. 
To minimize the impact of the mutations on the original optimal path $\pi(s^*)$, 
for any ODS $\hat{s}=\{\mathcal{A}, \mathcal{P}\}$ derived from the seed ODS $s^*$, 
we initially calculate a set of non-invasive feasible areas.
% Given the current ODS $\hat{s}=\{\mathcal{A}, \mathcal{P}\}$ derived from the seed ODS $s^*$, 
These areas are designated so that the introduction of a new participant $P_m$ to $\hat{s}$ is unlikely to influence the optimal \route $\pi(s^*)$
and other participants $\mathcal{P}$ in $\hat{s}$ if $P_m$ navigates within these non-invasive zones.
Note that
% based on the non-invasive mutations, the newly added participants $\mathcal{P}$ are unlikely to influence the original optimal path, and 
reducing the influence on the motions of other participants is also important, as their behaviors could indirectly influence the driving path of the ego vehicle.
% The constraint of not affecting the original participants is crucial to guarantee that other participants will also not influence the motion of the ego vehicle in the mutated scenarios.

The determination of the feasible area at any given moment~($t$) must account for the effects of both time and space. Specifically, it involves ensuring that the ego vehicle's original optimal \route $\pi(s^*)$ and the {\route}s of $\mathcal{P}$
% he relevant objects near the optimal path, including the ego vehicle, 
remain unaffected during a certain time frame ($\Delta t$) surrounding that moment.
% Given the current ODS $\hat{s}=\{\mathcal{A}, \mathcal{P}\}$ derived from the initial ODS $s$, and 
Suppose the current observation is $O(\hat{s})=\{\hat{o}_0, \hat{o}_1, \ldots, \hat{o}_k\}$ with a sampling time step $\Delta t$, we define the non-invasive feasible area regarding to the object $p\in \{\mathcal{P}\cup P_0\}$ during $[t,t+\Delta t]$ as $R_p(\hat{s}, t, \Delta t)$, where $P_0$ is the ego vehicle. 
% We substitute the ego observation in $O(\hat{s})$ with that in $O(s^*)$ to guarantee that the optimal path $\pi(s^*)$ remains unaffected.
% Given an ODS $s=\{\mathcal{A}, \mathcal{P}\}$ and the observation $O(s)=\{s_0, s_1, \ldots, s_k\}$ with a sampling time step $\Delta t$, we define the non-invasive feasible area regarding to the object $p\in \{\mathcal{P}\cup P_0\}$ during $[t,t+\Delta t]$ as $R_p(s,t,\Delta t)$, where $P_0$ is the ego vehicle.

\begin{figure}[!t]
    \centering
	\includegraphics[width=0.9\linewidth]{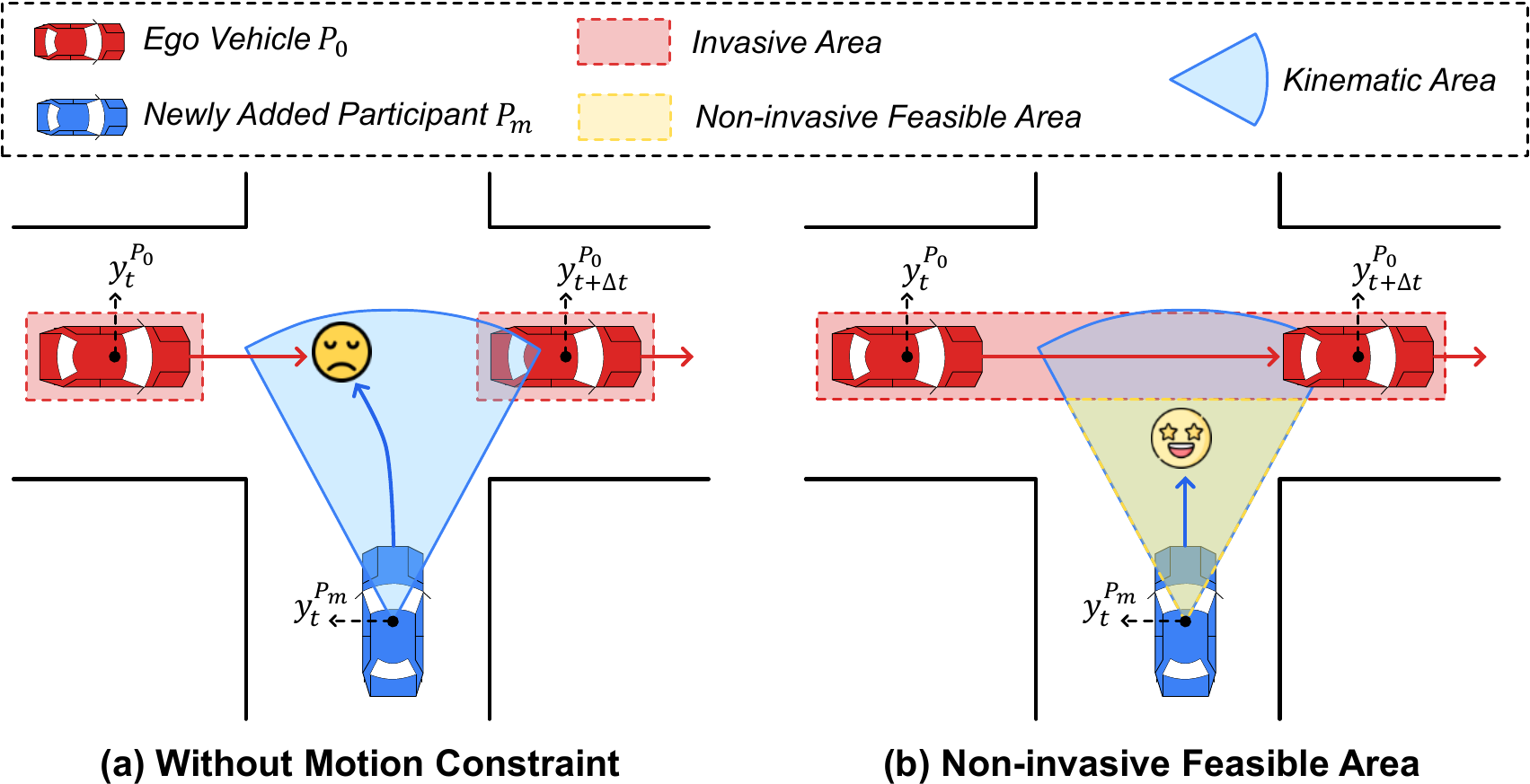} %decictor_mutation_v2
 \vspace{-5pt}
	\caption{Illustrative example of feasible area calculation.}
 	\label{fig:mutation}
  \vspace{-15pt}
\end{figure}
Fig. \ref{fig:mutation} illustrates the basic idea about the computation of $R_p(\hat{s},t,\Delta t)$, taking the ego vehicle as an example.
\cmf{We first estimate the feasible motion area of the new participant, denoted as $R(y_{t})$. Specifically, given the motion constraints (i.e., speed and steering ranges) and the position $y_{t}$ at timestamp $t$, we can calculate a sector range in Fig. \ref{fig:mutation}(a) according to the kinematic model~\cite{kong2015kinematic}. }
Second, we estimate the motion area of the ego vehicle $P_0$ between $[t, t+\Delta t]$ based on the original optimal \route $\pi(s^*)$, which is denoted as $R^{P_0}(\pi(s^*), \Delta t)$, e.g., the dashed light \cmf{red} area in Fig. \ref{fig:mutation}(b). \cmf{Note that the motion area of the ego vehicle is directly calculated from the real movement recorded in the scenario observations, which can be regarded as a rectangle between $t$ and $t+\Delta_{t}$.}
% $R^{P_0}(s_{t},s_{t+\Delta t})$ can be determined as $P_0$'s motion area from $s_{t}$ to $s_{t+\Delta t}$ in the {\ods} $s$.
Intuitively, the new participant should not move into $R^{P_0}(\pi(s^*), \Delta t)$ during the time frame $[t, t+\Delta t]$ so as not to affect the optimal \route of the ego vehicle.

Therefore, we can calculate the non-invasive feasible area (e.g., the light \cmf{yellow} area in Fig. \ref{fig:mutation}(b)) as:
\begin{equation}
    R_{P_0}(\hat{s},t,\Delta t)=R(y_{t}) \setminus R^{P_0}(\pi(s^*), \Delta t),
\end{equation}
% $R_{P_0}(s,t,\Delta t)=R(y_{t}) \setminus R^{P_0}(s_{t},s_{t+\Delta t})$, 

Similarly, we can calculate the non-invasive feasible areas $R_p(\hat{s}, t, \Delta t)$ with respect to any participant $p\in \mathcal{P}$ based on its driving \route $\pi^p(\hat{s})$.
The inference of the non-invasive feasible area at timestamp $t+\Delta t$ should consider all newly added participants, which is calculated as:
\begin{equation}
    R(\hat{s}, t, \Delta t) =\cap_{p\in \{\mathcal{P}\cup P_0\}} R_{p}(\hat{s}, t, \Delta t).
\end{equation}
% where $R_p(s,t,\Delta t)$ is the non-invasive feasible area regarding to the participant $p$, and $P_0$ is the ego vehicle.

Note that the time step $\Delta t$ will affect the computation of the non-invasive area. As $\Delta t$ increases, the calculation of the non-invasive area becomes more conservative as the new participant does not affect the ego behavior in longer time.

\begin{algorithm}[!t]
\small
\SetKwInOut{Input}{Input}
\SetKwInOut{Output}{Output}
\SetKwInOut{Para}{Parameters}
\SetKwFunction{area}{\textbf{NonInvasiveArea}}
\SetKwFunction{random}{\textbf{RandomNoise}}
\SetKwFunction{sample}{\textbf{Sample}}
\SetKwFunction{Add}{{\textbf{Adding}}}
\SetKwFunction{Modify}{{\textbf{Modifying}}}
\SetKwFunction{Remove}{{\textbf{Removing}}}
\SetKwFunction{Change}{{\textbf{Changing}}}
\SetKwFunction{Mutate}{{\textbf{Mutation}}}
\SetKwFunction{Oper}{{\textbf{Operator}}}
\SetKwProg{Fn}{Function}{:}{}

\Input{Participants in the seed ODS $s^*$: $\mathcal{P}_{s^*}$ \\
The ego vehicle's initial optimal \route: $\pi(s^*)$\\
% \cmf{Observation of the ego in $s$: $O({s})=\{{s}_{0}, {s}_{1}, \ldots, {s}_{k}\}$} \\
Participants in the current ODS $\hat{s}$: $\mathcal{P}_{\hat{s}}$ \\
Observation of $\hat{s}$: $O(\hat{s})=\{\hat{o}_{0}, \hat{o}_{1}, \ldots, \hat{o}_{k}\}$
% \todo{Add initial ego observation?}
}
\Output{Participants in the newly mutated scenario $s'$:  $\mathcal{P}_{s'}$
    % Mutated scenario participants $\mathcal{P}_{s'}$
}
\Para{Time step $\Delta t$ between two successive waypoints. 
}

\Fn{\Add{}}{
    $P_m \gets \{\}$\\ 
    $y_{0} \gets$ \textit{a collision-free waypoint} \\
    % \For{$s_{i} \in \{s_{1}, s_{2}, \ldots, s_{k}\}$}{
    \For{$t \in \{0,1,2,\ldots, k-1\}$}{
        % $\mathbf{R}^{i} \gets$ \area{$y_{i-1}, \{s_{i-\delta}, \ldots, s_{i+\delta}\}, \delta$} \\
        $R(\hat{s}, t, \Delta t)\gets$ \area{$y_{t}, \pi(s^*), O(\hat{s}), \Delta t$}\\
        \If{$R(\hat{s}, t, \Delta t)$ is $\emptyset$}{
            \Return $\emptyset$ \\
        }
        $y_{t+\Delta t} \gets$ \sample{$R(\hat{s}, t, \Delta t)$} \\
        $P_{m} \gets P_{m} \cup \{y_{t + \Delta t}\}$
    }
    $\mathcal{P}_{s'} \gets \mathcal{P}_{\hat{s}} \cup \{{P_{m}}\}$ \\
    \Return $\mathcal{P}_{s'}$
}

\Fn{\Remove{}}{
    $P_m \gets $ \sample{$\mathcal{P}_{\hat{s}} \setminus \mathcal{P}_{s^*}$} \\
    $\mathcal{P}_{s'} \gets \{\mathcal{P}_{\hat{s}} \setminus P_{m}\}$ \\
    \Return $\mathcal{P}_{s'}$
}

\cmf{
\Fn{\Change{}}{
    $\mathcal{P}_{s'} \gets \Remove{}$ \\
    $O(\hat{s}) \gets O(\hat{s}) \setminus \mathbf{y}^{\{\mathcal{P}_{\hat{s}} \setminus \mathcal{P}_{s'}\}}$ \\
    $\mathcal{P}_{\hat{s}} \gets \mathcal{P}_{s'}$ \\
    $\mathcal{P}_{s'} \gets \Add{}$ \\
    \Return $\mathcal{P}_{s'}$
}
}

\caption{Non-invasive Mutation Operators}
\label{algo-mutation}
\end{algorithm}

\textbf{Mutation Operations}.
Algorithm \ref{algo-mutation} outlines the specific mutation operations: \textit{Adding} (Lines 1-11), \textit{Removing} (Lines 12-15) \cmf{and \textit{Changing} (Line 16-21)}. The non-invasive mutation takes as input $\mathcal{P}_{s^*}$ (the participants in the seed ODS $s^*$), $\mathcal{P}_{\hat{s}}$ (the participants in the current ODS $\hat{s}$) and $O(\hat{s})$ (the observation of $\hat{s}$). The observation consists of a sequence of scenes $\{\hat{s}_0,\ldots,\hat{s}_k\}$ with an equal time step $\Delta t$. In each iteration, the non-invasive mutation randomly chooses one operation to change participants in the scenario $\hat{s}$, resulting in a new scenario~$s'$.

\textit{Adding.} 
Adding operation aims to introduce complexity to the scenario by adding a new participant to $\hat{s}$ that is less likely to influence the optimal path. In detail, this operation initializes an empty waypoint set $P_m$ and an initial collision-free waypoint $y_{0}$ (Lines 2-3). Waypoints for the added participant are iteratively generated (Lines 4-9), where each iteration involves calculating the non-invasive area (Line 5), sampling a non-invasive waypoint (Line 6-8), and adding it to $P_{m}$ (Line 9). The process concludes by merging the resultant set with the original participants $\mathcal{P}_{\hat{s}}$ to form $\mathcal{P}_{s'}$ (Line 10-11).

% \cmf{
% (2) \textit{Modifying.} This operation adds noise to existing participants in scenario $s$. It starts by selecting a random participant in $s$ but not in $\hat{s}$ and initializing an empty waypoint set $P'_{m}$ (Line 14). Unlike the adding operator, it applies random perturbations $\Delta_{i}$ to waypoints (Lines 17-21), adding perturbed waypoints to $P'_{m}$ if they are feasible (Lines 18-19), or retaining the original waypoint otherwise (Lines 20-21). The mutated participant set $\mathcal{P}{s'}$ is formed by replacing $P_m$ with $P'_{m}$ (Lines 22-23).
% }
% \todo{Delete modifying?}

\textit{Removing.} Continuously adding new participants to the scenario will introduce too many obstacles, resulting in motion task failures.
% thereby blocking the ego vehicle if the ego chooses a non-optimal path. 
Thus, we introduce the removing operator. Note that only the participants in $\mathcal{P}_{\hat{s}}\setminus \mathcal{P}_{s^*}$, i.e., the participants newly added to the seed ODS $s^*$, can be removed, while the participants in the seed ODS will remain unchanged in all mutated scenarios.
Because the removal of the participants in the seed ODS may affect the optimal path of the scenarios.
Specifically, this operation removes a participant in $\mathcal{P}_{\hat{s}}\setminus \mathcal{P}_{s^*}$ randomly (Lines 13-14) and returns the mutated set (Line 15).
% Note that the removed participant is not part of the initial ODS, as its removal could affect the optimality of paths in the mutated scenario.

\cmf{\textit{Changing.} The changing operation aims to modify the scenario $\hat{s}$ by replacing an existing participant in $\mathcal{P}_{\hat{s}}$. This operation involves first removing an existing object from $\mathcal{P}{\hat{s}}$ (Line 17) and then adding a new one (Line 20). Specifically, we remove the observation of the removed participant $\mathbf{y}^{\{\mathcal{P}_{\hat{s}}}$ from $O(\hat{s})$ (Lines 18-19) to ensure that the calculation of the non-invasive feasible area does not include this participant.}

\subsection{Consistency Check $\epsilon$}
\begin{figure}
    \centering
    \includegraphics[width=.98\linewidth]{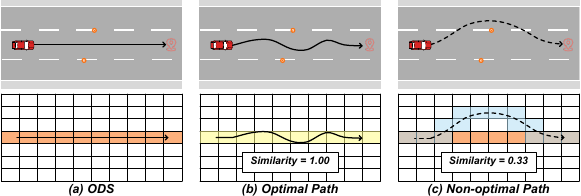}
    \vspace{-5pt}
    \caption{\cmf{Consistency Checking. \textbf{Above:} Driving paths. \textbf{Below:} Locations of the corresponding paths in the grid map.}}
    \label{fig:mr-check}
    \vspace{-15pt}
\end{figure}
Given the seed ODS $s^*$ and a mutated scenario $s'$, we require a criterion $\epsilon$ to determine whether $s'$ remains the optimal \route in $s^*$. A direct approach might be checking if the driving \route in $s'$ (i.e., $\tau(s')$) and the original \route of $s^*$ (i.e.,  $\tau(s^*)$) are the same. However, \cmf{even in real-world scenarios, human drivers do not strictly adhere to a straight line but exhibit some degree of zig-zag motion, as shown in Fig.~\ref{fig:mr-check}(b). This motion can still be considered optimal even if it slightly deviates from the original \route $\tau(s^*)$.} Thus, we propose an abstraction-based method to quantify the similarity of the two driving paths.

% it can be overly strict, as $\tau(s')$ could still be considered optimal even if it slightly deviates from the original \route $\tau(s^*)$, \cmf{such as the light degree of zig-zag motion in Fig.~\ref{fig:mr-check}(b).} Thus, we propose an abstraction-based method to quantify the similarity of the two driving paths.

% \fei{add figure: 3 sub figures}
Fig.~\ref{fig:mr-check} illustrates the basic idea of our method. We define an optimal area around the optimal \route $\tau(s^*)$ in terms of the grid map. Any driving \route falling or mostly falling within this area is considered optimal. 
For example, the \cmf{orange} area in Fig.~\ref{fig:mr-check}(a) defines the optimal area of the driving path in the seed \ods. \cmf{The driving path in Fig.~\ref{fig:mr-check}(b) does not strictly adhere to the optimal path but remains within the optimal area, achieving the highest similarity and thus meeting the optimality criterion. In contrast,} most of the path in Fig.~\ref{fig:mr-check}(c) (i.e., the \cmf{blue} area) falls outside the optimal area, resulting in a low similarity and a non-optimal path.

Specifically, the comparison between the driving {\route}s $\tau(s^*)$ and $\tau(s')$ is implemented by a grid-based approach, as shown in Fig.~\ref{fig:mr-check}. The map is divided into grids, and each driving \route is mapped to a set of grids. 
Technically, a driving \route of the ego vehicle in a scenario $s$ is represented as a sequence of locations $\tau(s) = \{p_0^0, \ldots, p_k^0\}$, where $p_i^0$ is the position at frame $i$. To collect the covered grids, a function $g$ is used to map each position $p$ to the grid where it is located. The covered grids for the \route $\tau(s)$ can be denoted as:
\begin{equation}\label{eq:gridcheck}
C_{\tau(s)} = \{{g}(p) \,|\,p\in \tau(s)\}
\end{equation}

To check the decision consistency, the similarity between the covered grids of $\tau(s^*)$ and $\tau(s')$ is computed. The consistency checking is defined using a predefined threshold $\varepsilon$ as follows:
\begin{equation}\label{eq:mrcheck}
\epsilon(\tau(s^*), \tau(s')) = \frac{|C_{\tau(s^*)} \cap C_{\tau(s')}|}{|C_{\tau(s^*)} \cup C_{\tau(s')}|} > \varepsilon
\end{equation}
If the similarity between the covered grids is greater than the threshold $\varepsilon$, $s'$ is recognized as an ODS.

\subsection{Feedback $\mathcal{D}$}\label{sec:fitness}
To guide the search of NoDSs, a feedback is necessary to select high-quality individuals from the candidate population (Line~\ref{algo0:select} of Algorithm~\ref{algo-metafuzz}). The grid similarity $\epsilon(\tau(s^*)$, $\tau(s'))$ (in Equation~\ref{eq:mrcheck}) provides a direct choice.
% for the feedback. 
However, the calculation of grid similarity is relatively coarse-grained as it only considers the spatial perspective of the ego vehicle's motion. 
While this coarse-grained calculation is beneficial for consistency checking, it may not be as specific and effective for guiding the testing to generate {\nods}s (see our evaluation results in Section~\ref{sec:feedbackperformance}). 
% Fine-grained feedback tends to provide more precise guidance for the testing process. 
To mitigate this, another metric is proposed that incorporates more fine-grained feedback \cmf{calculated from scenario observations (detailed in Section \ref{def-scenario}}), taking into account the ego vehicle's driving \route and motion behavior.
% driving \route and the behavior of the ego vehicle. 
The former focuses on the spatial characteristics of the ego vehicle's motion, while the latter more on the temporal ones.

\textit{Driving Path Feedback.} 
The main objective of \tool is to maximize the difference between the driving paths such that the non-optimal decision is identified. Considering the potential different lengths of $\tau(s^*)$ and $\tau(s')$, we use the average point-wise  distance between $\tau(s^*)$ and $\tau(s')$ in the spatial space to measure their difference, which is calculated as:
\begin{equation}
    f_{p}(s^*, s') = \frac{1}{n_{s'}} \sum_{i=1}^{n_{s'}} \min\{\|p_{i}-p\|\,|\, \forall p\in \tau(s^*)\}
    % f_{p}(s, s') = \min(\{\|p-p'\|| \forall p\in \tau(s) \wedge \forall p'\in \tau(s')\})
    % \sum_{i=0}^{|T_{s}|}min\{\|p_{i}^{0}, p_{k}^{0}\||0 \leq k \leq|T_{s_{0}}|\}.
\end{equation}
where $p_{i} \in \tau(s')$ and $n_{s'}$ is the total number of points in $\tau(s')$.

\textit{Behavior Feedback.} 
In some cases, optimizing only the driving \route feedback is difficult, thus an additional feedback mechanism is provided, which is based on the fine-grained behavior of the ego vehicle. This behavior feedback considers factors such as the ego's velocity, acceleration, and heading. By analyzing the ego's behavior, it becomes possible to gain insights into how slight changes in behavior could lead to variations in the driving path and increase the overall difference between $\tau(s^*)$ and $\tau(s')$. For example, if the heading of the ego vehicle changes slightly, it may not directly cause a significant increase in driving path differences. However, such a change could serve as a valuable indicator, as altering the heading or acceleration of the ego vehicle could lead to adjustments in the driving path, potentially resulting in an increased difference between paths.

To implement this behavior feedback, the ego's behavior is collected from its waypoints in the \cmf{scenario observation $O(s)$}, denoted as $\mathbf{X}_s = ((\theta_0, v_0, a_0), (\theta_1, v_1, a_1), \ldots, (\theta_k, v_k, a_k))$, where $\theta_i$, $v_i$ and $a_i$ represent the heading, velocity and acceleration at timestamp $i$, respectively.
To compare the behavior differences between two scenarios $s^*$ and $s'$, the Maximum Mean Discrepancy (MMD) is used as the measure of distance between their behavior distributions. The MMD is a widely used statistical metric for comparing distributions and can effectively quantify differences between two sets of data.
The behavior feedback function $f_b(s^*, s')$ is defined as:
\begin{equation}
f_b(s^*, s') = f_{MMD}(\mathbf{X}_{s^*}, \mathbf{X}_{s'}).
\end{equation}

Finally, our fitness score is calculated as:
\begin{equation}
    \mathcal{D}(\tau(s^*), \tau(s')) = f_{p}(s^*, s') + f_{b}(s^*, s').
\end{equation} 

\section{EMPIRICAL EVALUATION}

In this section, we aim to empirically evaluate the capability of \tool on NoDS generation. In particular, we will answer the following research questions:

\noindent \textbf{RQ1:} Can \tool effectively find NoDSs for ADSs in comparison to the baselines?

\noindent \textbf{RQ2:} How useful are the Non-invasive Mutation and the Feedback designed in \tool?

\noindent \textbf{RQ3:} How does \tool perform from the perspective of time efficiency?

%How is the performance of the \tool? 

To answer these research questions, we conduct experiments using the following settings:

\noindent \textbf{Environment.} We implement \tool on Baidu Apollo 7.0~\cite{apollo} and its built-in simulation environment SimControl.
Baidu Apollo 7.0 is an open-source and industrial-level ADS that supports a wide variety of driving supports. 
% Our experiments run on a Ubuntu 20.04 server with AMD EPYC 7543P, NIVIDIA RTX A5000 24GB and 256GB memory.

\noindent \textbf{Driving Scenarios.} We evaluate \tool on a real-world Map Sunnyvale Loop, provided as part of Baidu Apollo. Following the existing works~\cite{av_fuzzer, cheng2023behavexplor}, we use \cmf{six} representative scenarios and build the initial {\ods}s, the inputs of Algorithm \ref{algo-metafuzz}. \cmf{These scenarios have been widely tested in previous literature~\cite{av_fuzzer, cheng2023behavexplor,thorn2018framework,zhou2023specification,icse_samota,huai2023doppelganger,huai2023sceno,css_drivefuzzer,lu2022learning} and have covered all possible road types in the Apollo map library.} The selected {\ods}s include left turn (\textit{S1}), right turn (\textit{S2}), lane following (\textit{S3}), U-turn (\textit{S4}), \cmf{crossing intersection (\textit{S5}), and existing driving road (\textit{S6}).} The optimality of seed {\ods}s is ensured through manual verification.

% The optimality of seed {\ods}s is ensured from three perspectives: (1) All authors manually analyze each scenario to confirm that the path is efficient and optimal for task completion. (2) One industrial co-author verifies compliance with traffic regulations and ensures there are no safety concerns. 
% (3) We check and confirm that the new paths in the generated NoDSs are less optimal than the paths in the seed {\ods}s.

% (3) We will further check and confirm that the new paths in the generated NoDSs are less optimal than the paths in the seed {\ods}s.
% Note that, to expose the path-planning decision robustness issues, it is not necessary for the seed {\ods}s to be the most optimal. Rather, it is enough if the new path in the generated NoDSs is less optimal than the path in the seed {\ods}s.

\noindent \textbf{Baselines.}
\cmf{Since this is the first work to specifically evaluate the optimality of PPDs, and there are no baselines that are directly related to PPD testing. Thus, we design two random strategies for fair comparison with our \tool, i.e., \randombasline and \randomM.} Specifically, 1) \randombasline, which does not have any feedback and mutation constraints. It randomly generates scenarios by adding or removing dynamic vehicles or static obstacles; 2) \randomM, which does not have feedback but uses our non-invasive mutation $\delta$. \cmf{Besides, we selected seven safety-related baselines for comparison: AVFuzzer~\cite{av_fuzzer}, SAMOTA~\cite{icse_samota}, BehAVExplor~\cite{cheng2023behavexplor}, DriveFuzz~\cite{css_drivefuzzer}, scenoRITA~\cite{huai2023sceno}, DoppelTest~\cite{huai2023doppelganger} and DeepCollision~\cite{lu2022learning}.} We understand that comparisons with the seven baselines, primarily designed for identifying safety-critical violations, may not be absolutely fair. However, \cmf{our empirical results demonstrate that they can still generate scenarios with non-optimal PPDs, which would be overlooked without a consistency check.} Thus, the comparison with these tools still underscore the need to test the optimality of \cmf{PPDs}.

\cmf{Note that not all of baselines were originally evaluated in the same simulation environment as \tool (i.e., SimControl + Apollo). Specifically, AVFuzzer, BehAVExplor, and DeepCollision were implemented based on the LGSVL simulator~\cite{rong2020lgsvl}, which has been sunsetted~\cite{sunset}. SAMOTA and DriveFuzz are developed for Pylot~\cite{gog2021pylot} and Autoware~\cite{autoware}, respectively, both running with the CARLA simulator~\cite{dosovitskiy2017carla}. 
Therefore, we invested significant effort in migrating them to our simulation environment for comparison. Since their core algorithms (e.g., surrogate models, seed selection, and feedback) are often general, we kept them the same as the original implementation. We mainly modified the simulation interface (changing the simulator APIs) so that they can run on SimControl and Apollo. Furthermore, we set the ego task consistent with our approach to facilitate comparison. Details are available on~\cite{ourweb}.} 

\noindent \textbf{Metrics.}
To facilitate comparison with the baseline techniques, we have incorporated our consistency check into them to gather the generated NoDSs. 
In our experiments, we utilize the metrics \#NoDS 
and \#NoDS-Hum to assess the effectiveness of NoDS generation. 
% \#NoDS is the number of NoDSs generated by different tools, and \#NoDS-Hum is the number of NoDSs that are further validated by humans. 
\#NoDS quantifies the number of potential NoDSs, identified through our consistency checking with a predetermined threshold $\varepsilon$. 

As discussed in Section~\ref{sec:nimutation}, although our mutation strategy is conservative, it is still impossible to guarantee with absolute certainty that mutations will not influence ego behavior. Additionally, the ADS may plan an alternative route that, from a human perspective, could also be deemed an optimal decision. This bears resemblance to the generation of invalid inputs in existing adversarial attacks~\cite{cw2017} or deep learning testing methodologies~\cite{dola2021distribution}, despite there are some mutation constraints. Considering these factors, we introduce \#NoDS-Hum, representing the number of NoDSs validated by human evaluation. \cmf{Specifically, the co-authors verify whether the new paths in NoDSs are truly less optimal than the paths in the seed {\ods}s.} A NoDS is only counted in \#NoDS-Hum if \textit{all} authors unanimously recognize it as non-optimal, indicating a clear-cut case of a non-optimal decision. 
 
Moreover, we also assess the success rate of non-invasive mutations, denoted as \%Mutation. Specifically, the evaluation involves replaying the ego vehicle in the mutated scenario alongside the optimal path of the ODS and verifying whether the motion task can be completed on this path.

\noindent \textbf{{Implementation}.} Following AVFuzzer \cite{av_fuzzer}, we set the population size $N$ in \tool to $4$. 
In our mutation, we consider both dynamic vehicles and static obstacles (i.e., traffic cones). According to our preliminary study~\cite{ourweb_study}, we set the time step $\Delta t$ in the non-invasive mutation to 2 seconds and the equality threshold in the Consistency Check to $\varepsilon = 0.6$. The grid size is set to 2 meters to recognize when the entire vehicle deviates from the optimal path. Similar to previous works \cite{icse_samota, haq2023many}, we repeat each experiment 10 times to mitigate the influence of the non-determinism inherent in the simulation-based execution of the ADS. For each run, the budget is set as four hours.
% Note that the non-determinism primarily comes from the parallel execution of different modules in the ADS, while the simulator is often stable.}

% , as we found that it was long enough to compare.
% based on our preliminary evaluation. 

\subsection{RQ1: Effectiveness of \tool}

\begin{table*}[!t]
\centering
  \begin{minipage}{0.85\textwidth}
    \caption{\cmf{Comparison results with baselines.}}
    \small
    \vspace{-5pt}
    \resizebox{\linewidth}{!}{
        \begin{tabular}{l|ccccccc|ccccccc}
        % \specialrule{0em}{0pt}{0pt}
        \toprule
        \multirow{2}*{\textbf{Method}} & \multicolumn{7}{c|}{\textbf{\#NoDS} (\textbf{\#NoDS-Hum})$\,\uparrow$ } & \multicolumn{7}{c}{\textbf{\%Mutation}$\,\uparrow$}  \\
        \cmidrule(lr){2-8}\cmidrule(lr){9-15} 
         & \textit{S1} & \textit{S2} & \textit{S3} & \textit{S4} & \textit{S5} & \textit{S6} & \cellcolor{lightgray!20}{\textit{Sum}} & \textit{S1} & \textit{S2} & \textit{S3} & \textit{S4} & \textit{S5} & \textit{S6} & \cellcolor{lightgray!20}\textit{Avg.} \\
         
        \midrule
        \textbf{AVFuzzer}  & 0.6 (0.3) & 1.5 (1.0) & 0.0 (0.0) & 1.2 (0.5) & 1.4 (0.9) & 5.0 (0.6) & \cellcolor{lightgray!20}9.7 (3.3) & 41.5 & 40.5 & 70.6 & 10.6 & 6.7 & 77.5 & \cellcolor{lightgray!20}{41.2}\\
        
        \textbf{SAMOTA}  & 1.5 (0.9) & 0.4 (0.3) & 0.0 (0.0) & 3.3 (1.2) & 7.2 (6.2) & 4.0 (0.8) & \cellcolor{lightgray!20}16.4 (9.4) & 58.6 & 50.1 & 49.6 & 55.7 & 54.1 & 68.7 & \cellcolor{lightgray!20}{56.1}\\
        
        \textbf{BehAVExplor}  & 2.5 (1.7) & 1.6 (0.7) & 0.0 (0.0) & 6.3 (2.3) & 3.8 (3.3) & 3.0 (0.4) & \cellcolor{lightgray!20}17.2 (8.4) & 29.7 & 36.3 & 36.6 & 32.1 & 16.4 & 40.8 & \cellcolor{lightgray!20}{32.0}\\

        \textbf{DriveFuzz}  & 0.1 (0.0) & 5.8 (2.6) & 0.0 (0.0) & 4.2 (1.2) & 1.5 (0.9) & 4.0 (0.5) & \cellcolor{lightgray!20}15.6 (5.2) & 23.8 & 55.5 & 41.2 & 34.7 & 17.1 & 58.5 & \cellcolor{lightgray!20}38.5 \\

        % I repeat two new times to replace the abnormal point in scenoRITA s5
        \textbf{scenoRITA}  & 0.7 (0.3) & 0.0 (0.0) & 0.0 (0.0) & 0.9 (0.2) & 5.2 (2.7) & 2.2 (0.4) & \cellcolor{lightgray!20}9.0 (3.6) & 92.4 & 94.7 & 98.2 & 94.7 & 96.1 & 96.9 & \cellcolor{lightgray!20}95.5 \\

        \textbf{DoppelTest}  & 0.0 (0.0) & 0.0 (0.0) & 0.0 (0.0) & 1.8 (0.6) & 0.0 (0.0) & 0.0 (0.0) & \cellcolor{lightgray!20}1.8 (0.6) & 94.7 & 79.1 & 97.9 & 97.6 & 96.1 & 97.6 & \cellcolor{lightgray!20}93.8\\

        \textbf{DeepCollision} & 2.0 (1.0) & 6.1 (1.8) & 0.0 (0.0) & 3.2 (1.0) & 2.1 (0.8) & 1.9 (0.2) & \cellcolor{lightgray!20}15.3 (4.8) & 58.0 & 65.5 & 44.0 & 48.4 & 54.9 & 62.0 & \cellcolor{lightgray!20}55.5 \\ 
        
        \midrule
        \textbf{\randombasline}  & 0.3 (0.0) & 3.0 (1.5) & 0.2 (0.2) & 1.9 (0.7) & 7.6 (6.8) & 3.7 (0.3) & \cellcolor{lightgray!20}{16.7 (9.5)} & 73.2 & 73.1 & 70.2 & 76.7 & 65.3 & 66.0 & \cellcolor{lightgray!20}{70.8} \\
        
        \textbf{\randomM}  &3.0 (1.4) & 5.1 (4.6) & 4.4 (4.4) & 7.2 (2.0) & 10.5 (8.1) & 5.2 (0.4) & \cellcolor{lightgray!20}35.4 (20.9) & 96.4 & 97.4 & 99.0 & 96.0 & 95.6 & 98.1 & \cellcolor{lightgray!20}97.1 \\
        
        \textbf{\tool}  & \textbf{9.4 (3.9)} & \textbf{11.9 (8.6)} & \textbf{9.0 (9.0)} & \textbf{15.7 (6.7)} & \textbf{12.1 (9.4)} & \textbf{5.8 (1.1)} & \cellcolor{lightgray!20}\textbf{63.9 (38.7)} & \textbf{98.0} & \textbf{97.5} & \textbf{99.4} & \textbf{97.5} & \textbf{96.6} & \textbf{98.2} & \cellcolor{lightgray!20}\textbf{97.9} \\
        
        \bottomrule
        \end{tabular}
    }
    % \caption{Results of the total number of unique violations.}
    \label{tab:RQ1-baseline}
  \end{minipage}%
  \vspace{-15pt}
\end{table*}

\begin{figure}[!t]
    \centering
	\includegraphics[width=1.0\linewidth]{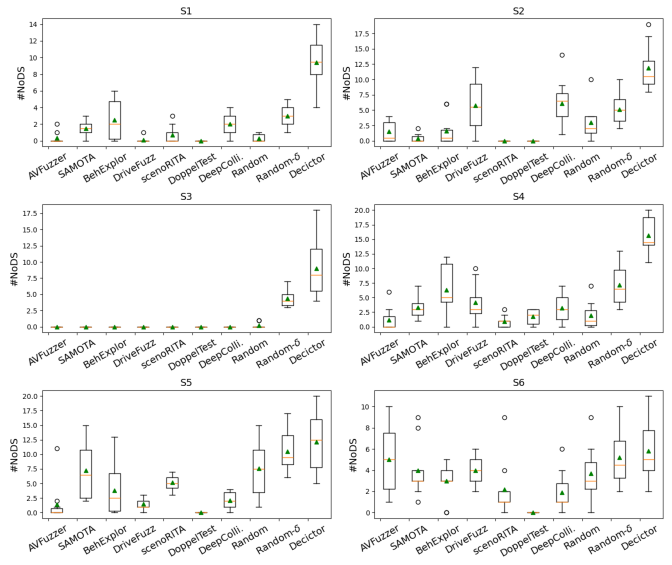}
 \vspace{-20pt}
	\caption{\cmf{Comparison of \#NoDS for different tools.}}
 	\label{fig:rq-1-nods}
  \vspace{-15pt}
\end{figure}

\subsubsection{Comparative Results} 
Table \ref{tab:RQ1-baseline} compares \#NoDS, \#NoDS-Hum, and \%Mutation across \cmf{six} initial {\ods}s: \textit{S1}, \textit{S2}, \textit{S3}, \textit{S4}, \cmf{\textit{S5}} and \cmf{\textit{S6}}, and Fig. \ref{fig:rq-1-nods} illustrates the distribution of \#NoDS across ten iterations of each method in every scenario, where the median and the average are represented by an orange bar and a green triangle, respectively.
From the results, we can find that \tool outperforms the baselines in \#NoDS and \#NoDS-Hum. 
In detail, on average of \#NoDS, \tool outperforms the best baseline (i.e., \randomM): \textit{S1} (9.4 vs. 3.0), \textit{S2} (11.9 vs. 5.1), \textit{S3} (9.0 vs. 4.4), \textit{S4} (15.7 vs. 7.2), \cmf{\textit{S5} (12.1 vs. 10.5) and \textit{S6} (5.8 vs. 5.2)}. 
Among all the detected NoDSs by \tool, we finally identified 3.9, 8.6, 9.0, 6.7, \cmf{9.4 and 1.1} NoDS-Hum in \textit{S1}, \textit{S2}, \textit{S3}, \textit{S4}, \cmf{\textit{S5} and \textit{S6}} respectively. 
{The detailed human verification results for NoDS-Hum are given on our website \cite{ourweb}.}
Even with very rigorous manual filtering, \tool still significantly outperforms the best baseline (\randomM).
 
The results indicate the effectiveness of \tool in identifying NoDSs and NoDS-Hum.
We further observed that \randomM outperforms both the \randombasline (\cmf{35.4 vs. 16.7} for the total number of NoDSs) and \cmf{the best safety-oriented baseline BehAVExplor (35.4 vs. 17.2 in \textit{Sum} of \#NoDS)}. 
This is attributed to the effectiveness of the non-invasive mutation used in \randomM and \tool. 
It is worth noting that BehAVExplor performs better than the other safety-guided baselines due to its diversity feedback mechanism (with diverse explorations), which makes it more likely to generate NoDSs.
\cmf{However, scenoRITA and DoppelTest exhibit lower performance.
% even though their \%Mutation are high  h. 
This occurs because they search for mutations in a wider road area than others, resulting in many mutations that do not impact the ego vehicle's motion. 
% across a broader range of map regions than others. 
scenoRITA outperforms DoppelTest as it incorporates a larger number of obstacles in its mutation processes, increasing the probability of generating NoDSs. We also observed that all safety-oriented baselines fail to perform on S3. This is primarily due to their lack of consideration for the impact of static obstacles during mutation, which is the key reason for NoDSs in S3.} 
% {}
\#NoDS-Hum is lower than \#NoDS due to the very strict criteria set for determining optimality. 

% focused on enhancing environmental and ego vehicle interactions for safety issue generation.
Regarding \%Mutation (\textit{Avg.} column), Table \ref{tab:RQ1-baseline} reveals that \randombasline and safety-oriented baselines produce a maximum of \cmf{95.5\% valid mutations (i.e., the original optimal path is not affected), which is lower than non-invasive mutation methods (97.1\% for \randomM and 97.9\% for \tool on average). The high rate of valid mutations demonstrates that our non-invasive mutation effectively ensures the validity of mutated objects.
% by utilizing an estimated safe and feasible area for modifying scenarios. 
Note that compared with other safety-oriented baselines, scenoRITA and DoppelTest generate a higher number of valid mutations. This is because they explore a broader mutation space, most of which is disjoint from the motion space of the ego vehicle,
% far from the within the map, 
which is less likely to conflict with the optimal path of the ego vehicle.}
% The comparison of results between \randomM and \tool highlights the effectiveness of the feedback mechanism in \tool. 
\cmf{Compared to \randomM, the average 0.8\% increase of \tool in \%Mutation indicates that integrating the feedback mechanism not only maintained the effectiveness of our mutation process but also increased the detection of NoDSs.
% after integrating the feedback mechanism, our mutation process maintained its effectiveness and the feedback increased the detected NoDSs.
}
% \cmf{The scenoRITA performs better than the other safety-guided baselines due to its high \%Mutation which makes it more likely to generate NoDSs. }
% BehAVExplor performs better than the other two safety-guided baselines due to its diversity feedback mechanism (with diverse explorations), which makes it more likely to generate NoDSs.

In summary, the comparison of \tool with \randombasline and \randomM demonstrates its effectiveness in detecting NoDSs. When contrasted with the \cmf{seven} state-of-the-art baselines, the results emphasize the importance of robust \cmf{path-planning decision} testing. \tool and these safety-oriented baselines are complementary, given their distinct testing objectives.

\begin{figure*}[!t]
    \centering
	\includegraphics[width=0.85\linewidth]{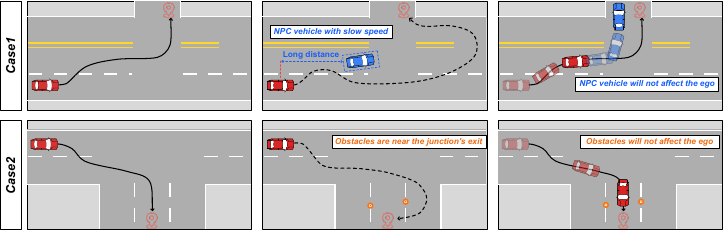}
\vspace{-5pt}	
 \caption{Cases of NoDSs detected by Decictor. The red vehicle is the AV. Optimal paths are indicated \cmf{\textbf{Solid lines}}, while non-optimal paths are displayed in \cmf{\textbf{Dotted lines}}. The first column is the initial ODSs, the second column shows the corresponding NoDSs, and the last column depicts the reproduced scenarios validating that the original optimal paths can be traversed in the NoDSs.}
 	\label{fig:rq-1-case}
  \vspace{-15pt}
\end{figure*}

{
\subsubsection{\cmf{Root Causes of NoDSs}} 
\cmf{To better understand the detected NoDSs, we manually summarize six patterns based on the potential root causes for these non-optimal decisions. Table~\ref{tab:RQ1-weight} shows the six unique root causes: }
\begin{enumerate}[leftmargin=*]
    \item \cmf{The ADS inaccurately predicts a moving NPC vehicle as stationary, leading to re-plan a non-optimal path.}
    \item \cmf{The ADS inaccurately enlarge the impact region of static obstacles on the ego vehicle's lane boundaries, resulting in the change of the optimal path.}
    \item \cmf{The ADS inaccurately estimates accessible regions between parallel obstacles that do not obstruct the optimal path.} 
    \item \cmf{The ADS overly prioritizes an unrelated NPC's movements, reducing feasible space for optimal path decision.}
    \item \cmf{The ADS overly focuses on keeping a safe distance from surrounding static obstacles, resulting in insufficient space to select the optimal path.}
    \item \cmf{The ADS is overly cautious about trailing vehicles, misclassifying part of the optimal area as non-optimal, resulting in insufficient space for executing the optimal path.}
\end{enumerate}
}

% {(1) The ADS inaccurately predicts a moving NPC vehicle as stationary, leading to re-plan a non-optimal path; (2) The ADS inaccurately enlarge the impact region of static obstacles on the ego vehicle's lane boundaries, resulting in the change of the optimal path;
% % identifies the impact of static obstacles on the roadside; 
% (3) The ADS inaccurately estimates the accessible region between parallel obstacles that do not obstruct the optimal path; (4) The ADS overly focuses on the movements of an unrelated NPC, resulting in insufficient feasible space to generate the optimal path;
% % causing it to overlook the inadequate space available for selecting the optimal path; 
% (5) The ADS overly focuses on keeping a safe distance from surrounding static obstacles, resulting in insufficient space to select the optimal path; (6) The ADS is excessively cautious about vehicles behind it and mistakenly considers part of the optimal area as non-optimal, leading to insufficient space for executing the optimal path.}

%persists in basing its decisions on the short-term traffic environment for an extended period, which restricts the timeframe available for choosing the optimal path; and 

\begin{table}[!t]
\centering
  \begin{minipage}{\linewidth}
   \centering
    \caption{\cmf{Proportion of discovered NoDSs}}
    % \vspace{-2.0pt}
    \vspace{-5pt}
    \small
    % \vspace{-10pt}
    \resizebox{\linewidth}{!}{
        \begin{tabular}{c|l|ccc}
        % \specialrule{0em}{0pt}{0pt}
        \toprule
        {\textbf{No.}} & \textbf{Root Cause} & \text{Prop.(\%)} & \text{Uni.} & \text{Uni.$^*$} \\
        \midrule
        1 & Inaccurate prediction of vehicle status & 6 & \checkmark & \checkmark \\
        % \midrule
        2 & Inaccurate prediction of obstacle impact & 29 & \checkmark & \ding{55}\\
        % \midrule
        3 & Inaccurate estimation of accessible region & 28 & \checkmark & \ding{55}\\
        % \midrule
        4 & Inaccurate prediction of other vehicles' intentions & 15 & \checkmark & \checkmark \\
        % \midrule
        5 & Excessive attention to surrounding obstacles & 10 & \checkmark & \ding{55}\\
        % \midrule
        6 & Excessively cautious about nearby vehicles & 12 & \ding{55} & \ding{55}\\
        \bottomrule
        \end{tabular}
    }
    % \caption{Results of the total number of unique violations.}
    \label{tab:RQ1-weight}
  \end{minipage}
  \vspace{-15pt}
\end{table}

\cmf{Table \ref{tab:RQ1-weight} shows the distribution of NoDSs discovered by \tool, with the respective proportions for the six root causes being 6\%, 29\%, 28\%, 15\%, 10\%, and 12\%, demonstrating \tool's capability to identify a diverse range of root causes in NoDSs. The \text{Uni.} column underscores that \tool identifies five unique root causes (No. 1 to 5) compared to the seven existing safety-critical ADS testing baselines. Notably, our method detects two unique NoDSs (No. 1 and 4) not identified by the Random-based baseline, \textit{Random-$\delta$} and \textit{Random}, thus highlighting \tool's enhanced ability to discover varied root causes in NoDSs.}

% \cmf{\textbf{Inaccurate Prediction of Vehicle Status.} The ADS inaccurately predicts a moving NPC vehicle as stationary, leading to an non-optimal path-planning decision.}

% \cmf{\textbf{Inaccurate Prediction of Obstacle Impact.} The ADS inaccurately perceives the target lane as impassable due to non-intrusive obstacles.}

% \cmf{\textbf{Inaccurate Estimation of Accessible Region.} The ADS inaccurately estimates the accessible region because of parallel obstacles, which do not actually obstruct the optimal path.}

% \cmf{\textbf{Inaccurate Prediction of Unrelated Vehicle’s Intentions.} The ADS overly focuses on the movements of an unrelated NPC, causing it to overlook the inadequate space available for selecting the optimal path.}

% \cmf{\textbf{Excessive Focus on Short-term Traffic Environment.} The ADS persists in basing its decisions on the short-term traffic environment for an extended period, which restricts the timeframe available for choosing the optimal path.}

% \cmf{\textbf{Unflexible path-planning decisions.} The ADS typically responds to nearby slow-moving vehicles by choosing to accelerate, and this intensive focus leaves inadequate room to execute the optimal path.}

\subsubsection{Case Study}
% \todo{divide the violations into different levels.}
% \todo{metion replay in evaluation part}
Fig.~\ref{fig:rq-1-case} shows two representative NoDS examples. More NoDS examples and corresponding videos can be found on our website~\cite{ourweb}. 
The examples illustrate that the ego vehicle takes non-optimal paths in the detected NoDSs, even though the optimal ones are still available. 

\textbf{Case 1. Inaccurate Prediction of Vehicle Status.} 
% \tool introduces a new NPC vehicle with a low initial speed in the NoDS. Note that the NPC vehicle is far away from the ego vehicle. Even though it is positioned on the ego vehicle's optimal path, it will have moved away by the time the ego vehicle reaches this segment, i.e., the original optimal path remains available (as seen in the optimal \route in the reproduced scenario). However, the ADS incorrectly predicts the status of the NPC vehicle and interprets it as a stationary one. As a result, it stops the lane changing and selects a very inefficient \route.
% \tool adds a new NPC vehicle with a low initial speed in the NoDS. Despite being initially placed on the ego vehicle's optimal path and far away, the NPC vehicle will have moved by the time the ego vehicle reaches that segment. The original optimal path remains available, as shown in the reproduced scenario. However, the ADS makes an incorrect prediction, considering the NPC vehicle as stationary. Consequently, it stops lane changing and chooses a less efficient path.
The \tool adds a new NPC vehicle in the NoDS, initially slow and located at a distance to the ego vehicle. As shown in the third column of Fig.~\ref{fig:rq-1-case}, by the time the ego vehicle approaches, the NPC has moved, yet the original optimal path is still available. However, due to an incorrect prediction treating the NPC vehicle as stationary, the ADS stops lane changing and chooses a less efficient path.
% \textcolor{red}{However, as shown in the third column of Fig.~\ref{fig:rq-1-case}, when the ADS stopped lane changing, the NPC vehicle moved to the lane that does not intersect with the original optimal path of the ego vehicle.}

\textbf{Case 2. Inaccurate Prediction of Obstacle Impact.} \tool places two obstacles near the junction exit, not affecting the optimal path. However, the ego vehicle (i.e., 2.1-meter width) deems the target lane (i.e., 3.5-meter width) impassable, choosing a suboptimal path from an adjacent lane, which raises risks in real-world traffic scenarios.

\begin{center}
\fcolorbox{black}{gray!10}{\parbox{0.96\linewidth}{\textbf{Answer to RQ1}: \tool significantly outperforms existing methods in identifying NoDSs. The non-invasive mutation can accurately generate valid scenarios \cmf{(96.6\% - 99.4\%)}.
}}
\end{center}

\subsection{RQ2: Usefulness of Mutation and Feedback} \label{sec:ex-rq2}
We evaluated the key components of \tool, including the \textit{non-invasive mutation} and \textit{feedback} mechanism, by configuring a series of \tool variants.
% Due to the space limit, we only show the experimental results of \#NoDS and put all the corresponding \#NoDS-Hum in \cite{ourweb}.

\subsubsection{Mutation} \label{sec:exp-mutation}
For mutation, we compared \tool with its three variants: (1) \textit{w/o Cons} applies a random mutation instead of the non-invasive mutation in \tool, aiming to evaluate the effectiveness of the non-invasive mutation;
(2) \textit{w/o Mot} sets the time step to 0s in the calculation of non-invasive feasible areas (i.e., only considering the motion constraint at each timestamp), aiming to measure the influence of the motion constraint between two successive timestamps.
(3) \textit{w/o Rem}  uses only the adding operation in \tool under the constraint of non-invasive mutation, aiming to assess the effectiveness of the combination of the two mutation operations, i.e., adding and removing participants.
As shown in Table \ref{tab:RQ2-mutation}, we find that \textit{w/o Cons} generates the fewest valid mutations \cmf{(69.2\%)} and detects the smallest number of NoDSs \cmf{(25.8)}, underlying the importance of the non-invasive mutation.
%, which sets strict constraints to ensure the optimal \route remains unaffected. 
Comparing \textit{w/o Mot} with \tool, we observed that only considering the motion constraint at each timestamp is insufficient, and the continuous-time motion constraint plays a significant role in calculating the non-invasive feasible area. 
Our preliminary study revealed that \%Mutation increases while the \#NoDS decreases as the time step increases. The reason is that a long time step will yield scenarios with reduced interactivity between the ego vehicle and the added participant, thus exerting a lower impact on the \cmf{PPD} process of the ADS. 
% Our preliminary study showed that \%Mutation increases while \#NoDS decreases with longer time steps, as extended time steps reduce interactivity between the ego vehicle and added participants, diminishing their impact on the ADS's \cmf{PPD} process.
% Detailed results regarding the influence of the time step are available on our website \cite{ourweb}. 
Comparing \textit{w/o Rem} and \tool, we found that mutation with only the adding operation is more likely to induce more invalid mutations and a lower number of NoDSs. 
This is because the adding operation introduces too many obstacles, potentially leading to the failure of the motion task.
% This is because the adding operation brings too many obstacles, which may cause the failure of the motion task.
% block the path of the ego vehicle. 
Thus, the removing operation is also important for generating NoDSs.

\begin{table}[!t]
    \centering
    \begin{minipage}{\linewidth}
   \centering
    \caption{\cmf{Effectiveness of mutations}}
    \small
    \vspace{-5pt}
    \resizebox{\linewidth}{!}{
        \begin{tabular}{c|l|cccccccc}
        % \specialrule{0em}{0pt}{0pt}
        \toprule
        {\textbf{Metric}} & {\textbf{Method}} 
         & \textit{S1} & \textit{S2} & \textit{S3} & \textit{S4} & \textit{S5} & \textit{S6} & \cellcolor{lightgray!20}{\textit{Avg.}} & \cellcolor{lightgray!20}{\textit{Sum}}\\
         
        \midrule
        
        \multirow{4}*{\textbf{\%Mutation}$\,\uparrow$} & \textbf{w/o Cons} &71.7 & 71.3 & 74.8 & 65.5 & 67.0 & 64.7 & \cellcolor{lightgray!20}{69.2} & \cellcolor{lightgray!20}- \\ 
        
        & \textbf{w/o Mot} & 71.1 & 80.9 & 98.6 & 68.9 & 77.6 & 66.1 & \cellcolor{lightgray!20}{77.2} & \cellcolor{lightgray!20}- \\ 

        & \textbf{w/o Rem} & 97.5 & 97.3 & 99.4 & 97.2 & 91.8 & 94.0 & \cellcolor{lightgray!20}{96.2} & \cellcolor{lightgray!20}- \\
         
        & \textbf{\tool} & \textbf{98.0} & \textbf{97.5} & \textbf{99.4} & \textbf{97.5} & \textbf{96.6} & \textbf{98.2} &  \cellcolor{lightgray!20}\textbf{97.9} & \cellcolor{lightgray!20}- \\ 

        \midrule
        
        \multirow{4}*{\textbf{\#NoDS}$\,\uparrow$} & \textbf{w/o Cons} & 2.7 & 4.5 & 1.6 & 5.7 & 7.0 & 4.3 & \cellcolor{lightgray!20}- & \cellcolor{lightgray!20}25.8 \\
        
        & \textbf{w/o Mot} & 4.0 & 8.6 & 5.2 & 10.8 & 11.7 & 5.0 & \cellcolor{lightgray!20}- & \cellcolor{lightgray!20}45.3 \\
        
        & \textbf{w/o Rem} & 3.9 & 9.3 & 4.9 & 9.1 & 8.3 & 5.7 & \cellcolor{lightgray!20}- & \cellcolor{lightgray!20}{41.2} \\
        
        & \textbf{\tool} & {\textbf{9.4}} & \textbf{11.9} & \textbf{9.0} & \textbf{15.7} & \textbf{12.1} & \textbf{5.8} & \cellcolor{lightgray!20}- & \cellcolor{lightgray!20}\textbf{63.9}\\
        
        \bottomrule
        
        \end{tabular}
    }
    % \caption{Results of the total number of unique violations.}
    \vspace{-10pt}
    \label{tab:RQ2-mutation}
  \end{minipage}
\end{table}

\begin{table}[!t]
\centering
  \begin{minipage}{0.75\linewidth}
   \centering
    \caption{\cmf{Comparison of feedback}}
    % \vspace{-2.0pt}
    \vspace{-5pt}
    \small
    % \vspace{-10pt}
    \resizebox{\linewidth}{!}{
        \begin{tabular}{l|ccccccc}
        % \specialrule{0em}{0pt}{0pt}
        \toprule
        \multirow{2.5} * {\textbf{Method}} & \multicolumn{7}{c}{\textbf{\#NoDS}$\,\uparrow$} \\
        \cmidrule(lr){2-8}
        &  \textit{S1} & \textit{S2} & \textit{S3} & \textit{S4} & \textit{S5} & \textit{S6} & \cellcolor{lightgray!20}\textit{Sum}  \\
        \midrule
        \textbf{F-Random} & 3.0 & 5.1 & 4.4 & 7.2 & 10.5 & 5.2 & \cellcolor{lightgray!20}35.4 \\
        
        \textbf{F-Con} & 3.0 & 6.7 &1.2 & 7.5 & 11.7 & 4.9 & \cellcolor{lightgray!20}{35.0}\\
        
        \textbf{F-Path} & 4.3 & 10.8 & 5.0 & 10.6 & 12.1 & 5.4 & \cellcolor{lightgray!20}{48.2}\\
        
        \textbf{F-Behavior}  & 5.3 & 9.0 & 5.4 & 11.5 & 11.8 & 5.6 & \cellcolor{lightgray!20}{48.6}\\
        
        \textbf{\tool}  & \textbf{9.4} & \textbf{11.9} & \textbf{9.0} & \textbf{15.7} & \textbf{12.1} & \textbf{5.8} & \cellcolor{lightgray!20}\textbf{63.9} \\
        \bottomrule
        \end{tabular}
    }
    % \caption{Results of the total number of unique violations.}
    \label{tab:RQ2-fitness}
  \end{minipage}
  \vspace{-15pt}
\end{table}

\subsubsection{Feedback} \label{sec:feedbackperformance}
For feedback, we implemented four variants: (1) \textit{F-Random} replaces the feedback of \tool with a random selection (from $\textbf{Q}\cup \textbf{Q}'$) to evaluate the effectiveness of our feedback strategy. (2) \textit{F-Con} replaces the 
% \textit{driving path feedback} 
feedback of \tool with the grid similarity used in our consistency checking (see Equation~\ref{eq:mrcheck}) to compare the performance between using a coarse-grained grid distance and a fine-grained path distance. (3) \textit{F-Path} and (4) \textit{F-Behavior} consider only the \textit{driving path feedback} and the \textit{behavior feedback}, respectively, to evaluate the usefulness of either feedback type.

Table \ref{tab:RQ2-fitness} shows the experimental results of \#NoDSs.
The comparative results between \textit{F-Random} and \tool~\cmf{(35.4 vs. 63.9 in total)} illustrate the effectiveness of the feedback used in \tool. The results of \textit{F-Con} \cmf{(35.0)} indicates that the grid similarity is not an effective feedback metric, as it 
% is coarse-grained and 
may overlook some scenarios with a high probability of inducing NoDSs. From the ablation results of \textit{F-Path} and \textit{F-Behavior} (\cmf{48.2 and 48.6}, respectively), we see that both \textit{behavior feedback} and \textit{driving path feedback} are beneficial for detecting NoDSs.
Their combination achieves the best performance. 

Note that, we also evaluated the influence of different $\Delta t$ used in non-invasive mutation. Due to the space limit, the detailed experimental results can be found on our website \cite{ourweb}.

% \fei{Due to the space limit, we put all the corresponding \#NoDS-Hum results on our website?}

% \begin{table}[]
%     \centering
%     \caption{Comparison of different feedback.}
%     % \vspace{-2.0pt}
%     \small
%     \resizebox{0.3\linewidth}{!}{
%         \begin{tabular}{l|ccccc}
%         % \specialrule{0em}{0pt}{0pt}
%         \toprule
%         \multirow{2} * {\textbf{Method}} & \multicolumn{5}{c}{\textbf{\#NoDS}} \\
%         \cmidrule(lr){2-6}
%         &  \textit{S1} & \textit{S2} & \textit{S3} & \textit{S4} & \cellcolor{lightgray!20}\textit{Sum}  \\
%         \midrule
%         \textbf{F-Random} & 3.0 & 5.1 & 4.8 & 9.5 & \cellcolor{lightgray!20}22.4 \\
        
%         \textbf{F-Con} & 3.0 & 6.9 &1.4 & 8.4 & \cellcolor{lightgray!20}{19.7}\\
        
%         \textbf{F-Path} & 4.3 & 10.4 & 6.8 & 12.5 & \cellcolor{lightgray!20}{34.0}\\
        
%         \textbf{F-Behavior}  & 5.8 & 11.0 & 6.0 & 10.0 & \cellcolor{lightgray!20}{32.8}\\
%         \textbf{\tool}  & \textbf{9.4} & \textbf{11.9} & \textbf{9.0} & \textbf{15.7} & \textbf{46.0} \\
%         \bottomrule
%         \end{tabular}
%     }
%     % \caption{Results of the total number of unique violations.}
%     \label{tab:RQ2-fitness}
% \end{table}

% \vspace{-5pt}
\begin{center}
\fcolorbox{black}{gray!10}{\parbox{0.96\linewidth}{\textbf{Answer to RQ2}: Both the non-invasive mutation and the two types of feedback are useful for \tool to detect non-optimal decisions.}}
\end{center}
\subsection{RQ3: Test Efficiency of \tool}

\begin{figure}[!t]
    \centering
	\includegraphics[width=0.9\linewidth]{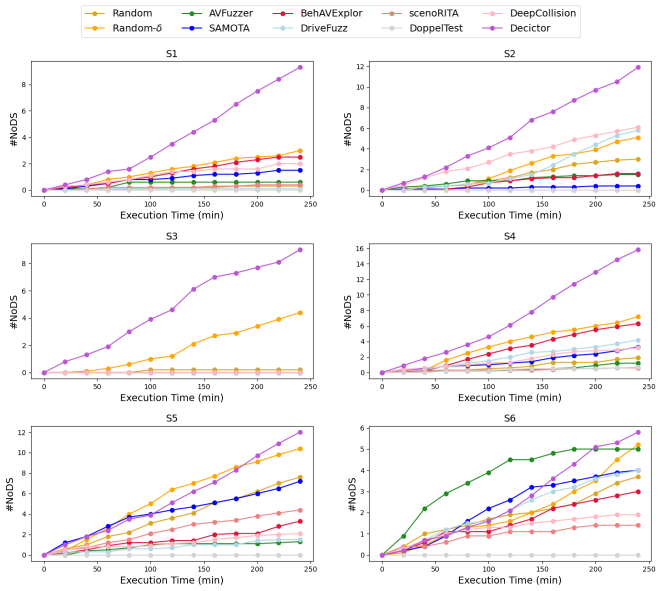}
 \vspace{-10pt}
	\caption{\#NoDSs over the execution of different methods.}
 	\label{fig:rq-3}
  \vspace{-5pt}
\end{figure}

\begin{table}[!t]
    \centering
    \caption{\cmf{Results of time performance (s)}}
    \vspace{-5pt}
    \small
    \resizebox{1.0\linewidth}{!}{
        \begin{tabular}{l|cccc|c}
        % \specialrule{0em}{0pt}{0pt}
        \toprule
         \textbf{Method} & \textbf{Mutation} & \textbf{Oracle Check} & \textbf{Feedback} & \textbf{Simulation} & \textbf{Total}  \\
        \midrule
        \textbf{Random} & 2.28 & N/A & N/A & 66.67 & 68.95 \\
        \textbf{Random-$\delta$} & 8.04 & N/A & N/A & 65.30 & 73.34\\
        \textbf{AVFuzzer} & 1.46 & 0.01* & 0.01* & 71.26 & 72.74 \\
        \textbf{SAMOTA} & 1.18 & 0.01* & 32.11 & 65.05 & 98.35 \\
        \textbf{BehAVExplor} & 0.56 & 0.01* & 1.41 & 77.06 & 79.04 \\
        \textbf{DriveFuzz} & 2.02 & 0.01* & 0.01* & 77.56 & 79.60 \\
        \textbf{scenoRITA} & 0.55 & 0.01* & 22.16 & 72.70 & 95.42 \\
        \textbf{DoppelTest} & 0.01 & 0.01* & 21.36 & 69.06 & 90.44 \\
        \textbf{DeepCollision} & 0.19 & 0.01* & 0.02 & 66.42 & 66.64 \\
        \midrule
        \textbf{Decictor} & 8.03 & 1.62 & 0.73 & 67.20 & 77.58 \\
        \bottomrule
        \end{tabular}
    }
    
    % \caption{Results of the total number of unique violations.}
    \label{tab:RQ4-performance}
    % \vspace{-2mm}
    \vspace{-15pt}
\end{table}

Fig. \ref{fig:rq-3} shows the cumulative number of NoDSs over the execution time of different methods. We can find that as \tool continues its execution, the detection of NoDSs steadily increases, whereas other methods quickly reach a stable state.
We further assess the time performance of different components in \tool, including the overhead of mutation, oracle checking, feedback calculation, and simulation. 
Specifically, we analyze the average time to handle a scenario.
The results are summarized in Table \ref{tab:RQ4-performance}, where the columns \textit{Mutation}, \textit{Oracle Check}, \textit{Feedback}, and \textit{Simulation} represent the average time taken by a scenario for mutation, validity checking, fitness computation, and running the scenario in the simulation platform, respectively. 0.01* represents that the time cost is small, i.e., less than 0.01.
For Random, the oracle and feedback stages are not involved, being marked as `N/A'. 

The overall results show that, for all tools, the simulation process spends the majority of time. 
For instance, on average, \tool takes 77.58 seconds to process a scenario, where 67.20 seconds (86.62\%) are used in running the scenario.
\tool spends slightly more time in Mutation and Oracle Check than others due to the computation of the non-invasive feasible area for each waypoint and the grid-based similarity-checking mechanism. \cmf{Note that we only consider the mutation time when generating scenario configurations.} 
However, the computation time of \tool remains within an acceptable range.
SAMOTA consumes the most time in the feedback due to the necessity of training a surrogate model. \cmf{The feedback times for scenoRITA and DoppelTest are approximately twice as long as those reported in~\cite{huai2023doppelganger}. This is due to the simulation length was set to twice the original settings to ensure the ego could complete the task.}
The differing time duration for simulation across various tools can be attributed to the different scenarios each tool generates (e.g., time-out scenarios), which in turn require varying amounts of simulation time.

% %[=]66.15273[=]                                          
% 2024-07-31 16:00:09.794 | INFO     | scenario.runner.scenario_runner_RL:_run_seed:215 - --> [Simulation Time] Mutation Spend Time: [=]0.20183100000000007[=]
% 2024-07-31 16:00:09.794 | INFO     | scenario.runner.scenario_runner_RL:_run_seed:216 - --> [Simulation Time] Feedback Spend Time: [=]0.007227000000000001[=]

% \begin{table}[!t]
%     \centering
%     \caption{Results of time performance (s)}
%     \vspace{-5pt}
%     \small
%     \resizebox{1.0\linewidth}{!}{
%         \begin{tabular}{l|cccc|c}
%         % \specialrule{0em}{0pt}{0pt}
%         \toprule
%          \textbf{Method} & \textbf{Mutation} & \textbf{Oracle Check} & \textbf{Feedback} & \textbf{Simulation} & \textbf{Total}  \\
%         \midrule
%         \textbf{Random} & 2.28 & N/A & N/A & 66.67 & 68.95 \\
%         \textbf{Random-$\delta$} & 8.04 & N/A & N/A & 65.30 & 73.34\\
%         \textbf{AVFuzzer} & 1.46 & 0.01* & 0.01* & 71.26 & 72.74 \\
%         \textbf{SAMOTA} & 1.18 & 0.01* & 32.11 & 65.05 & 98.35 \\
%         \textbf{BehAVExplor} & 0.56 & 0.01* & 1.41 & 77.06 & 79.04 \\
%         \textbf{Decictor} & 8.03 & 1.62 & 0.73 & 67.20 & 77.58 \\
%         \bottomrule
%         \end{tabular}
%     }
%     % \caption{Results of the total number of unique violations.}
%     \label{tab:RQ4-performance}
%     % \vspace{-2mm}
%     \vspace{-20pt}
% \end{table}

\vspace{-5pt}
\begin{center}
\fcolorbox{black}{gray!10}{\parbox{0.96\linewidth}{\textbf{Answer to RQ3:} \tool demonstrates efficiency, with the majority of the time (86.62\%) spent in the simulation phase, while the main algorithmic component consumes approximately 10.38 seconds (13.38\%).}}
\end{center}

\subsection{Threats to Validity}
\tool suffers from some threats. 
First, the selection of the seed {\ods}s could potentially influence the results. To mitigate this, \cmf{we have judiciously selected six motion tasks and created corresponding scenarios. The decision optimality of each {\ods} was manually analyzed and confirmed by all authors, including a co-author with industrial expertise. Second, the threshold selection for the Consistency Check $\epsilon$ in \tool presents another threat, as different thresholds can lead to varying interpretations of non-optimal path decisions. } 
%For instance, a large threshold might classify significant zig-zag motions as NoDSs.
% despite the Consistency Check $\epsilon$ in \tool, the manual confirmation of NoDS-Hum may present another threat as different people may identify a scenario as either an {\ods} or a {\nods}. 
To counter this, on one hand, we carefully select the threshold $\varepsilon$ in the Consistency Check and $\Delta t$ in the computation of non-invasive feasible area (the empirical evaluation of the two parameters can be found on our website \cite{ourweb}) to reduce the number of such scenarios; on the other hand, to ensure robustness and consistency, all co-authors independently review the generated {\nods}s, and only those confirmed by unanimous agreement among all authors will be considered in NoDS-Hum.
Third, \cmf{the non-determinism inherent in ADS execution may also pose a threat to our results,} and we mitigate it by repeating each experiment multiple times.

Lastly, the environments used could pose a potential threat. On one hand, we adapted existing baselines to the Apollo+SimControl simulation environment, which could potentially influence the results. However, our modifications were mainly limited to the interface between the algorithms and the simulation environment, leaving the core algorithms unchanged. We thoroughly reviewed the code and released the implementation of baselines on our website~\cite{ourweb}.

% Another potential threat could be that our current Decision Robustness primarily addresses non-optimal decisions related to selecting inefficient paths. However, there can be other types of non-optimal decisions, such as lack of smooth driving or unnecessary jittery movements, which are not considered. In the future, we plan to extend the Decision Robustness (e.g., considering velocity and acceleration) to detect more non-optimal decisions.
% \cmf{Another potential threat could be that our mutation operators are limited to dynamic vehicles and static obstacles (i.e., traffic cones). Although existing mutation operators have demonstrated good performance, introducing different mutation operators may influence the results. In the future, we plan to extend our tool to support more mutation operators, such as pedestrians.}
 
% On the other hand, reusing \tool on different simulators or ADSs may have varying performances. Similar to previous works, \tool can be adapted to other platforms by implementing an adaptor. We plan to adapt \tool to other environments in the future~work.}

% the customization of baselines to the Apollo+SimControl simulation environment could potentially affect the results. However, we only modified the interface between the algorithms and the simulation environment, leaving the algorithms themselves untouched. Moreover, to ensure the correctness, all co-authors thoroughly reviewed the code. \cmf{Reference implementation of all baselines is available on our website~\cite{ourweb}.}
 
\section{Related Work}

\noindent \textbf{Safety-Guided ADS Testing.}
Safety plays the most important role in the development of ADSs. 
However, guaranteeing everlasting safety is challenging and many approaches have been proposed to generate safety-critical scenarios for evaluating the safety of ADSs. 
They can be divided into data-driven approaches \cite{zhang2023building,deng2022scenario,gambi2019generating,najm2013depiction,nitsche2017pre,roesener2016scenario, paardekooper2019automatic,lu2024diavio}
and searching-based approaches \cite{cheng2023behavexplor,hildebrandt2023physcov,gambi2019automatically,han2021preliminary,av_fuzzer,icse_samota,tse_adfuzz,tang2021systematic,zhou2023specification,tang2021route,tang2021collision,huai2023doppelganger,li2023generative}. 
Data-driving methods produce critical scenarios from real-world data, such as traffic recordings~\cite{deng2022scenario,paardekooper2019automatic,roesener2016scenario,tang2023evoscenario} and accident reports~\cite{gambi2019generating,najm2013depiction,nitsche2017pre,zhang2023building,guo2024sovar,tang2024legend}.
Search-based methods use various technologies to search for safety-critical scenarios from the scenario space, such as \cmf{guided fuzzing~\cite{cheng2023behavexplor, av_fuzzer, MDPFuzz_2022_issta, tian2022generating}}, \cmf{evolutionary algorithms~\cite{gambi2019automatically,han2021preliminary, tang2021collision,tang2021route,tang2021systematic,zhou2023specification,tian2022mosat}}, metamorphic testing~\cite{han2020metamorphic}, surrogate models~\cite{icse_samota,tse_adfuzz}, \cmf{reinforcement learning \cite{haq2023many, feng2023dense, lu2022learning}} and reachability analysis~\cite{hildebrandt2023physcov,althoff2018automatic}.
% Metamorphic testing is another efficient tool to generate safety-critical scenarios~\cite{han2020metamorphic}. 
All these methods aim to generate critical scenarios for evaluating ADSs' safety-critical requirements, such as collision avoidance and reaching the destination.
In contrast, our work evaluates the non-safety-critical requirements of ADSs, which is also crucial for the real-world deployment of ADSs. \cmf{Compared to these works, the key difference is that \tool is the first to evaluate non-safety-critical requirements for ADSs, particularly in the aspect of non-optimal path-planning decisions, whereas the other works mainly focus on safety-critical issues.}

% \cmf{Compared to these works, the key difference is that \tool is the first that focuses on evaluating  non-safety-critical requirements for ADSs, particularly in the aspect of non-optimal path-planning decisions, while they mainly focus on the safety-critical issues.}

% Search-based safety testing \cite{cheng2023behavexplor}, \cite{hildebrandt2023physcov}

% Data-Driven safety testing \cite{zhang2023building}, \cite{deng2022scenario} proposed a test reduction and prioritization method to automatically extract and analyze rich driving scenarios from real-world driving records, such as accident reports and driving videos.

\noindent \textbf{Robustness Analysis for ADS.} 
Robustness is a crucial issue for autonomous driving systems (ADSs). Recent studies have shown that the perception module in ADSs (i.e., object detection) exhibit vulnerabilities to various threats including perturbing sensor signals \cite{li2021fooling, cao2019adversarial}, modifying objects \cite{gao2024multitest, gao2023benchmarking, cao2023you}, etc. 
% For example, the authors [M] inserted physical-aware objects to generate corner cases for evaluating multi-sensor fusion perception module in the ADS. 
There are also some studies \cite{zhang2022adversarial, cao2022advdo} work on the robustness of the prediction module by crafting trajectories of other vehicles. However, no existing work has tested the \cmf{path-planning} robustness of autonomous driving systems. In this paper, we propose the first work to reveal non-optimal \cmf{PPDs} of the ADS, which can be meaningful to inspiring the improvement on ADS robustness and reliability.

\noindent \cmf{\textbf{Path-planning in ADSs.} Obtaining optimal \cmf{PPDs} is a challenging task in ADSs~\cite{reda2024path}. Existing studies on path planning primarily focus on avoiding obstacles, overlooking other attributes that contribute to optimality (e.g., motion time, path length, and comfort). Traditional algorithms~\cite{warren1993fast,thoresen2021path,spanogiannopoulos2022sampling,li2022autonomous} often result in discontinuous paths, leading to non-optimal overall decisions. While, machine learning-based planning algorithms~\cite{jiang2022robust,viana2018distributed,zhang2020optimal} could become trapped in local minima, unable to find optimal solutions. 
Therefore, given an ADS, it is critical to evaluate the robustness of its optimal PPDs, but there is still a significant gap in this area.
To mitigate this, we proposed the first PPD testing technique.}

% Metamorphic testing (MT) has been an emerging approach to generating test cases and overcoming the oracle problem~\cite{chen2018metamorphic, xie2011testing}.
% Recently, MT has also been applied to test deep learning systems, such as the perception system of ADSs \cite{xie2022towards} and end-to-end deep learning-based ADSs~\cite{zhang2018deeproad, wang2020metamorphic,deng2022declarative}.
% For example, the authors in \cite{wang2020metamorphic} designed an MR to test the robustness of an object detection system, which can be roughly described as: Inserting
% an additional object that does not overlap with
% preexisting objects in the source image should not affect the detection of the preexisting objects in the new image. 
% However, there is only a little work focusing on MT for multi-module ADSs~\cite{zhou2019metamorphic,han2020metamorphic}.
% In \cite{han2020metamorphic}, the authors propose an MT method to identify avoidable collision scenarios, which means there are genuine failures in the ADS under test.  
% In contrast to existing MT techniques that primarily focus on the correctness and safety of the system under test, we propose an MT method to detect vital non-safety-critical violations of the ADS under test.   

\section{Conclusion}
\cmf{In this paper, we propose the first study to evaluate the robustness of ADSs' path-planning decisions.} To address this issue, we develop a testing tool \tool, which comprises three main components: 
Non-invasive Mutation, generating new scenarios that are unlikely to affect the original optimal paths; Consistency Check, determining whether the driving paths in the mutated scenarios are consistent with the original optimal one; Feedback-guided Selection, responsible for selecting better scenarios for the subsequent mutation.
The experimental results demonstrate the effectiveness and efficiency of \tool, as well as the usefulness of each component in \tool. 

\section*{Acknowledgment}

\noindent 
This research is partially supported by the Ministry of Education, Singapore under its Academic Research Fund Tier 2 (Proposal ID: T2EP20223-0043), the Lee Kong Chian Fellowship, the National Natural Science Foundation of China (62102283) and the Beijing Nova Program. Any opinions, findings and conclusions or recommendations expressed in
this material are those of the author(s) and do not reflect the views of the Ministry of Education, Singapore.

\bibliographystyle{IEEEtran}
\bibliography{reference}
% \bibliography{,reference.bib}{}

\end{document}